\documentclass[10pt,journal,compsoc]{IEEEtran}
\usepackage{hyperref}

\usepackage{verbatim}
\usepackage{xcolor}

\ifCLASSOPTIONcompsoc
  \usepackage[nocompress]{cite}
\else
  \usepackage{cite}
\fi
\ifCLASSINFOpdf
  \usepackage[pdftex]{graphicx}
\else
  \usepackage[dvips]{graphicx}
\fi
\usepackage{amsmath}
\ifCLASSOPTIONcompsoc
  \usepackage[caption=false,font=footnotesize,labelfont=sf,textfont=sf]{subfig}
\else
  \usepackage[caption=false,font=footnotesize]{subfig}
\fi
\begin{document}
%

\title{An Explicit Local and Global Representation Disentanglement Framework\\
with Applications in Deep Clustering and Unsupervised Object Detection
}
%
%
%
%


\author{Rujikorn Charakorn,
        Yuttapong Thawornwattana,
        Sirawaj Itthipuripat,
        Nick Pawlowski,
        Poramate Manoonpong,
        and~Nat Dilokthanakul
\IEEEcompsocitemizethanks{
\IEEEcompsocthanksitem N. Dilokthanakul, Y. Thawornwattana,
R. Charakorn and P. Manoonpong are with the school of information science and technology, Vidyasirimedhi Institute of Science and Technology (VISTEC), Rayong,
Thailand.
\IEEEcompsocthanksitem S. Itthipuripat is with Learning Institute and Futuristic Research in Enigmatic Aesthetics Knowledge
Laboratory, King Mongkut's University of Technology Thonburi (KMUTT), Bangkok, Thailand.
\IEEEcompsocthanksitem N. Pawlowski is with Imperial College London.
\IEEEcompsocthanksitem N. Dilokthanakul and R. Charakorn are the main contributors and contributes equally to this work.
\IEEEcompsocthanksitem Corresponding author: Nat Dilokthanakul  (E-mail: natd\_pro@vistec.ac.th)
}
\thanks{Manuscript received }}

%
%

\markboth{}%
{Shell \MakeLowercase{\textit{et al.}}: Bare Demo of IEEEtran.cls for Computer Society Journals}
%



\IEEEtitleabstractindextext{%
\begin{abstract}

Visual data can be understood at different levels of granularity, where global features correspond to semantic-level information and local features correspond to texture patterns. In this work, we propose a framework, called SPLIT, which allows us to disentangle local and global information into two separate sets of latent variables within the variational autoencoder (VAE) framework. Our framework adds generative assumption to the VAE by requiring a subset of the latent variables to generate an auxiliary set of observable data. This additional generative assumption primes the latent variables to local information and encourages the other latent variables to represent global information. We examine three different flavours of VAEs with different generative assumptions. We show that the framework can effectively disentangle local and global information within these models leads to improved representation, with better clustering and unsupervised object detection benchmarks. Finally, we establish connections between SPLIT and recent research in cognitive neuroscience regarding the disentanglement in human visual perception. The code for our experiments is at \url{https://github.com/51616/split-vae}.

\end{abstract}

}

\maketitle

\IEEEdisplaynontitleabstractindextext

%
\IEEEpeerreviewmaketitle

\IEEEraisesectionheading{\section{Introduction}\label{sec:introduction}}

\setlength{\textfloatsep}{10pt plus 1.0pt minus 2.0pt}
\IEEEPARstart{H}{umans} understand visual information at different levels of granularity. We understand high-level visual concepts and are also able to describe granular details of what we perceive. Our perceptual system can ignore irrelevant information (“invariant" property) and only attend to information that is useful for the task at hand ("selective" property) \cite{Rust2010selectivity}. This ability of attending to the most salient features is thought to help focus the computation on the most useful and relevant information, thus accelerate learning, aid generalisation, and save computational cost \cite{bengio2017consciousness}. For example, sensitivity to the variations in colour and lighting is, generally, undesirable. These variations should be ignored in some tasks, e.g. object detection, which depend more on the global or coarser context information. 

Deep learning models are hierarchical in their nature. 
They exhibit an ability to combine increasingly complex features into high-level outputs. 
However, the emergence of such hierarchical features is crucially dependent on the data and the optimisation dynamics during learning.
It has been observed that local features are regularly used for class prediction 
if there are unintended biases or spurious correlations in the dataset \cite{jo2017measuring,geirhos2018imagenettrained}. 
For example, if a deep neural network is trained to classify boats from cars. 
It would correctly guess that an object is a boat if it can detect water texture 
because it learns spurious correlations between the water texture and the boat label. 
We say that the representation is \emph{entangled} as the data is not represented in a more meaningful way.

One way to combat this problem is to have as many variations in the data 
as possible to help the models \emph{disentangle} less important variations or nuisance factors 
from the more important ones, which would allow for better generalisation across tasks and situations \cite{james2017transferring,tobin2017domain}.
However, this approach depends critically on the data collection process, which can be restrictive in some domains. Another family of methods performs the disentanglement \emph{explicitly} on the structure of the models \cite{eslami2016attend,shanahan2019explicitly}. While the emerged latent structure can be more restrictive, these explicit disentanglement methods have representations that are more predictable and robust because the imposed biases are more independent from the optimisation dynamics.  

In this work, we propose a representation learning mechanism 
that explicitly disentangles the global variations from the local ones. 
Motivated by the benefits of explicit disentangled representations, we enforce the local variations to be modelled separately. This separation has several benefits: For example, it is easy to ignore local variations by only using the representation of global variations for a downstream task. Moreover, additional representation structures can be imposed selectively on the global or local variations. 

Our proposed disentanglement method is an extension to the variational autoencoder (VAE) framework \cite{kingma2013auto,rezende2014stochastic}. A VAE is a deep generative model which learns a data generating process by assuming that the observed data are generated from a deep transformation of latent variables. In this framework, the latent variables are assumed to be independent factors which generate the observable data, i.e. these independent factors (coupled with the generation process) govern the variations in the data. Thus, the training of VAE forces the inferred representation to be independent as specified by the prior assumptions.

We can also add more complex structures to the representation by augmenting the generative model assumptions with the desired structures. For example, the Gaussian-Mixture VAE (GMVAE) \cite{dilokthanakul2016deep,shu2017note} assumes a multi-modal latent distribution, as opposed to a uni-modal Gaussian as commonly used in VAE, allowing it to cluster data. Another notable example is the Attend-Infer-Repeat (AIR) model \cite{eslami2016attend}. It creates the representation of an object’s position, size, identity and quantity by using a special generative mechanism that involves the sequential drawing of each object onto the scene. This structure allows AIR to perform unsupervised object detection and counting.  

\begin{figure*}[t]
    \centering
    \includegraphics[trim={0.5cm 17.5cm 0.5cm 0.5cm},clip, width=0.80\linewidth]{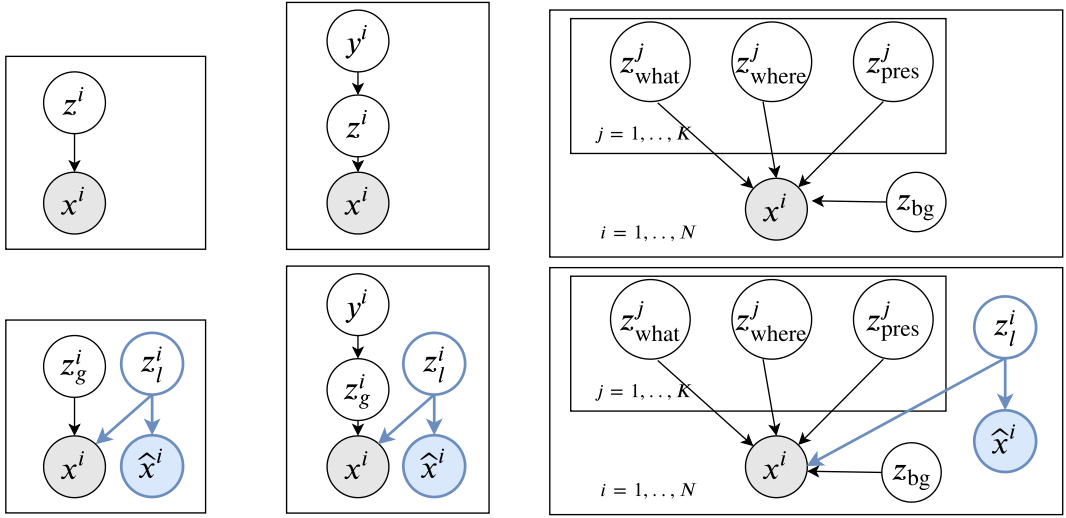}
    \caption{The Separated Paths for Local and Global Information (SPLIT) framework. VAE, GMVAE and SPAIR are three examples of deep generative models under the VAE framework. The left-most model is a vanilla VAE, where the latent variable ($z$) has a unit-variance Gaussian distribution. The middle model is the GMVAE, where latent structure is a multi-modal distribution. The right-most model is the SPAIR, where the data are assumed to be generated from structured latent variables such as the identity ($z_{\text{what}}$), the location ($z_{\text{where}}$), and the visibility ($z_{\text{pres}}$) [\textbf{TOP ROW}].  These generative models can be augmented with additional structures under the SPLIT framework. We create $\hat{x}$ from $x$ in such a way that it only contains local variation in $x$. This forces the model to represent the local variation in the additional latent variable $z_l$, and encourages the remaining latent variables to only represent global variation [\textbf{BOTTOM ROW}]. }
    \label{fig:graphical}
\end{figure*}

Within the VAE framework, we propose a special generative model assumption that allows the global variations of the data to be modelled separately from the local ones (Fig. \ref{fig:graphical}).
We call this \emph{\textbf{S}eparated \textbf{P}aths for \textbf{L}ocal and Global \textbf{I}nforma\textbf{t}ion} (SPLIT) framework. This is done by creating auxiliary observable data $\hat{x}$ by transforming the data $x$ using a random scrambling operation. This procedure removes global-scale correlations from $x$, leaving $\hat{x}$ with only local information. A set of latent variables that are responsible for the generation of $\hat{x}$ is then forced to model only the local variations. Importantly, the remaining latent variables can then focus its representation on the global variations of the data. 

The proposed method can be applied flexibly to a VAE model with any complex latent structure. 
We showed the usefulness of our approach by applying it to three different VAE models: (i) the original VAE \cite{kingma2013auto, rezende2014stochastic}, (ii) the Gaussian-Mixture VAE \cite{dilokthanakul2016deep, shu2017note} and (iii) the Spatially-Invariant AIR \cite{crawford2019spatially,eslami2016attend}. We then showed empirically that disentanglement of local and global variations can help improve the interpretability of the representations and also improve generalisation ability of the models. The main contributions of this paper are as follow: 
\begin{itemize}
    \item We propose a method for extending VAE models by allowing them to explicitly disentangle global and local variations in the data.\footnote{The methodological contributions in this work is based on a chapter of N.D.'s PhD thesis \cite{nat2019}, which have not been published as a paper.}
    \item The proposed method is generic and can be integrated seamlessly within the VAE framework. 
    We also show empirically that our method can be flexibly applied to VAE models with rich latent assumptions. 
    \item We show that deep unsupervised clustering can be performed with explicit bias imposed,  resulting in more interpretable representations.
    \item We show that our method improves an unsupervised object detection algorithm by making it more robust against background variations.
\end{itemize}

\section{Background}

The proposed method in this paper is an extension to the variational autoencoder (VAE) framework \cite{kingma2013auto,rezende2014stochastic}.
In this section, we give a brief review of latent variable models and the VAE. 
We also describe the Gaussian-Mixture VAE (GMVAE) and the Spatially-Invariant AIR (SPAIR)~\cite{crawford2019spatially}. Readers familiar with the VAE and its extensions can safely skip this section.

Latent variable models are a class of models that 
describes the observed data in terms of relationships
between observable variables and latent variables. 
Data is assumed to be 
distributed according to a generative process where latent variables generate observations through conditional transformations.
These latent variables
are believed to be the sources of variation in the observed data, and
lie in more compact, lower dimensional spaces than the data space. 
Latent factors that capture meaningful features or variations of the data constitute 
a good representation of the data. 
Bayesian inference can be used to infer these latent factors from
the data, revealing disentangled representations.

\subsection{Variational Autoencoder}
Variational Autoencoder (VAE) is a latent variable model whose
conditional distribution of the observed variable given the latent variables is parameterised by a deep neural network transformation \cite{rezende2014stochastic,kingma2013auto}. 
More specifically, the data are assumed to come from
the following generative process:
\begin{align}
    z &\sim p(z), \\
    x &\sim p_{\theta}(x|z),
\end{align}
where $z$ is the latent variable, $x$ is the observable variable 
and $\theta$ are parameters of the deep neural network. 

VAE learns $\theta$ by maximising the evidence lower-bound (ELBO),
i.e. a lower bound on the log probability of 
the observations. 
For each datapoint, the ELBO is:
\begin{align}
    &\log p_{\theta}(x^{i}) \nonumber\\
    \geq &L_{ELBO}(\theta, \phi, x^{i}) \nonumber \\
    = &E_{q_{\phi}(z|x^{i})}[\log p_{\theta}(x^{i}|z)] - D_{KL}(q_{\phi}(z|x^{i})| p(z) ),
\end{align}
where $q_{\phi}$ is the variational posterior which is another deep neural network parameterised by $\phi$ and $p(z)$ is the prior distribution. 
The embedding distribution $q_{\phi}(z)$ is encouraged to be distributed as $p(z)$.
The deep neural network parameters $\theta$ and $\phi$ can be optimised using the standard stochastic gradient descent and backpropagation algorithm.

\subsection{Gaussian Mixture Variational Autoencoder}
\label{sec:gmvae}

The original VAE has a unit-variance Gaussian as the prior distribution $p(z)$.
This choice is used because it is mathematically simple and is easy to implement. 
Because this prior has a simple isotropic structure, the encoded representation 
has a limited representational power \cite{MathRainSiddTeh2019}.
In some cases, it might be more useful to model the data with a multi-modal 
representation or other complex structures. 

VAEs, when used with a multi-modal prior, 
can represent data that has cluster structure \cite{dilokthanakul2016deep, shu2017note}. For example, 
using a mixture of Gaussians as the prior distribution for the VAE results in 
a clustering model. We refer to this VAE model with a multi-Gaussian 
latent prior as the Gaussian-Mixture VAE (GMVAE).

An example of a GMVAE's generative process is
\begin{align}
    y &\sim p(y), \\
    z &\sim p_{\gamma}(z|y), \\
    x &\sim p_{\theta}(x|z),
\end{align}
where $\gamma$ and $\theta$ are two sets of deep neural network parameters, $p(y)$ is a discrete distribution and $p_{\gamma}(z|y)$ is a Gaussian condition on $y$. Marginally, $p_{\gamma}(z)$ has a Gaussian mixture distribution with $k$ components, 
\begin{align}
    p_{\gamma}(z) = \sum_{j=1}^{k} p_{\gamma}(z|y = j) p(y = j).
\end{align}

Optimising the ELBO can be more complicated for this type of mixture priors than the standard VAE. 
As the VAE uses stochastic gradient descent with backpropagation, the computation 
leading to the evaluation of ELBO needs to be differentiable. To avoid 
non-differentiable paths in the computation, a relaxation trick is required. 
In this work, we use the Gumbel-Softmax distribution \cite{jang2016categorical, maddison2016concrete}
instead of the discrete distribution.

\subsection{Spatially Invariant Attend Infer Repeat}

In addition to the multi-modal structure, one might want to encode a more complex structure 
into the representation. Attend-Infer-Repeat (AIR) \cite{eslami2016attend} is a VAE model that 
explicitly represents objects in image data in terms of their position, size, identity and the number of times they appear. 
This richly structured representation allows the model to count and detect objects in 
an unsupervised manner. 

In order for AIR to represent these structures, it leverages a deep neural network 
architecture called the spatial transformer network (STN) \cite{jaderberg2015spatial}. STNs work by transforming an 
input image to a specific location and scale, representing a glimpse of the image. Importantly, a STN transformation 
is learnable and end-to-end differentiable. By using a STN as an encoder and decoder, AIR can reason about locations and scales of objects.

In order for AIR to count the number of objects, it uses an recurrent neural network (RNN) \cite{greff2016lstm}
along with an adaptive computation mechanism. 
The RNN in AIR can decide whether to stop its encoding-decoding computations. 
By assuming that each encoding-decoding step corresponds to one object, 
AIR can count the number of objects through the length of the RNN computation.

Spatially invariant AIR (SPAIR) \cite{crawford2019spatially} is an extension to AIR. 
Similar to AIR, it encodes objects' locations and identity. 
But unlike AIR, it uses a convolution neural network instead of an RNN to encode and decode objects, partly simultaneously, making it much more scalable and robust. Fig. \ref{fig:graphical} shows the graphical models of the original VAE, GMVAE and SPAIR. 
These models make different generative assumptions, resulting in different representation structures.

\section{SPLIT Framework}

\begin{figure}[!tb]
    \centering
    \subfloat[VAE\label{1a}]{%
       \includegraphics[width=0.8\linewidth, trim={0 5.5cm 0 7cm}, clip]{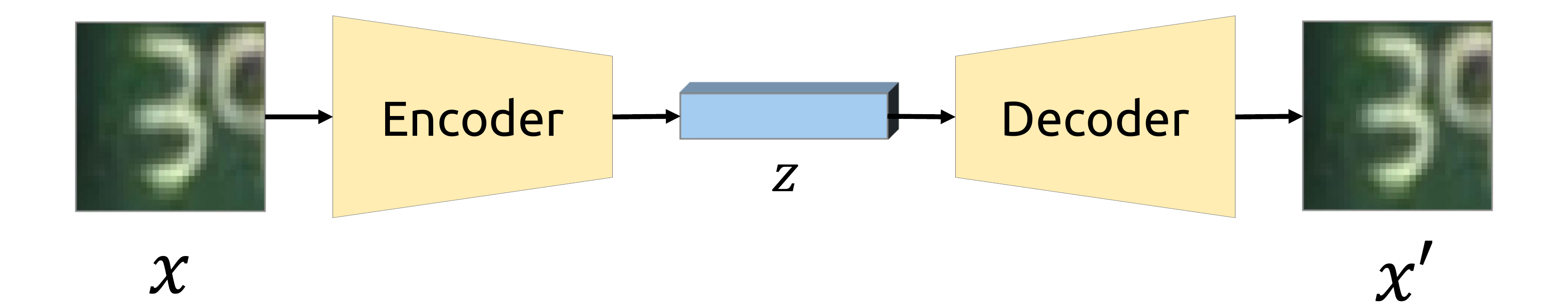}}
       \\
    \subfloat[SPLIT-VAE\label{1b}]{%
        \includegraphics[width=0.8\linewidth, trim={0 2.5cm 0 2.5cm}, clip]{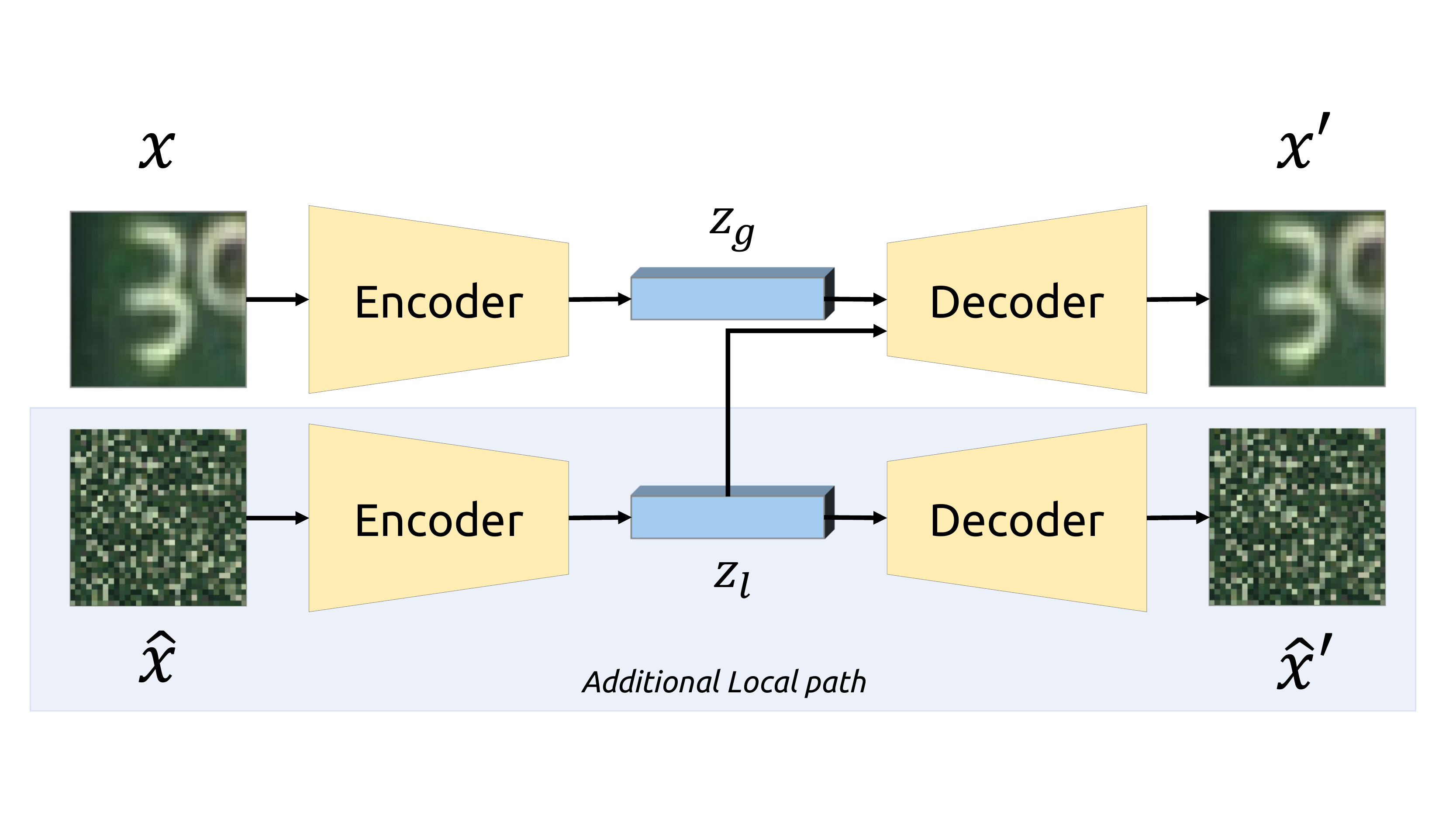}}
       \\
    \subfloat[SPLIT-GMVAE\label{1c}]{%
        \includegraphics[width=0.8\linewidth, trim={0 0cm 0 0.8cm}, clip]{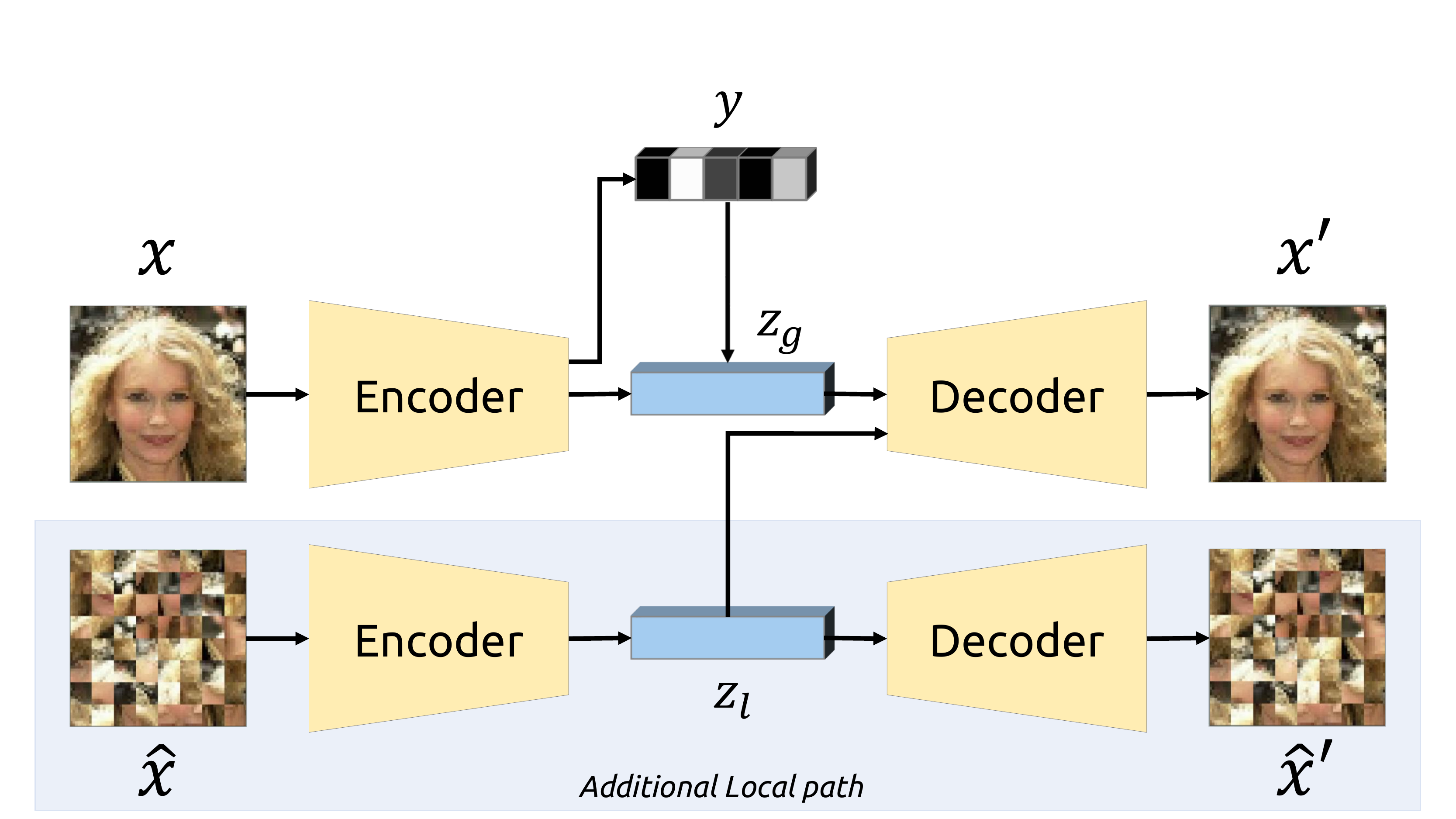}} 
       \\
    \subfloat[SPLIT-SPAIR\label{1d}]{%
        \includegraphics[width=1.0\linewidth, trim={0 2.33cm 0 0cm}, clip]{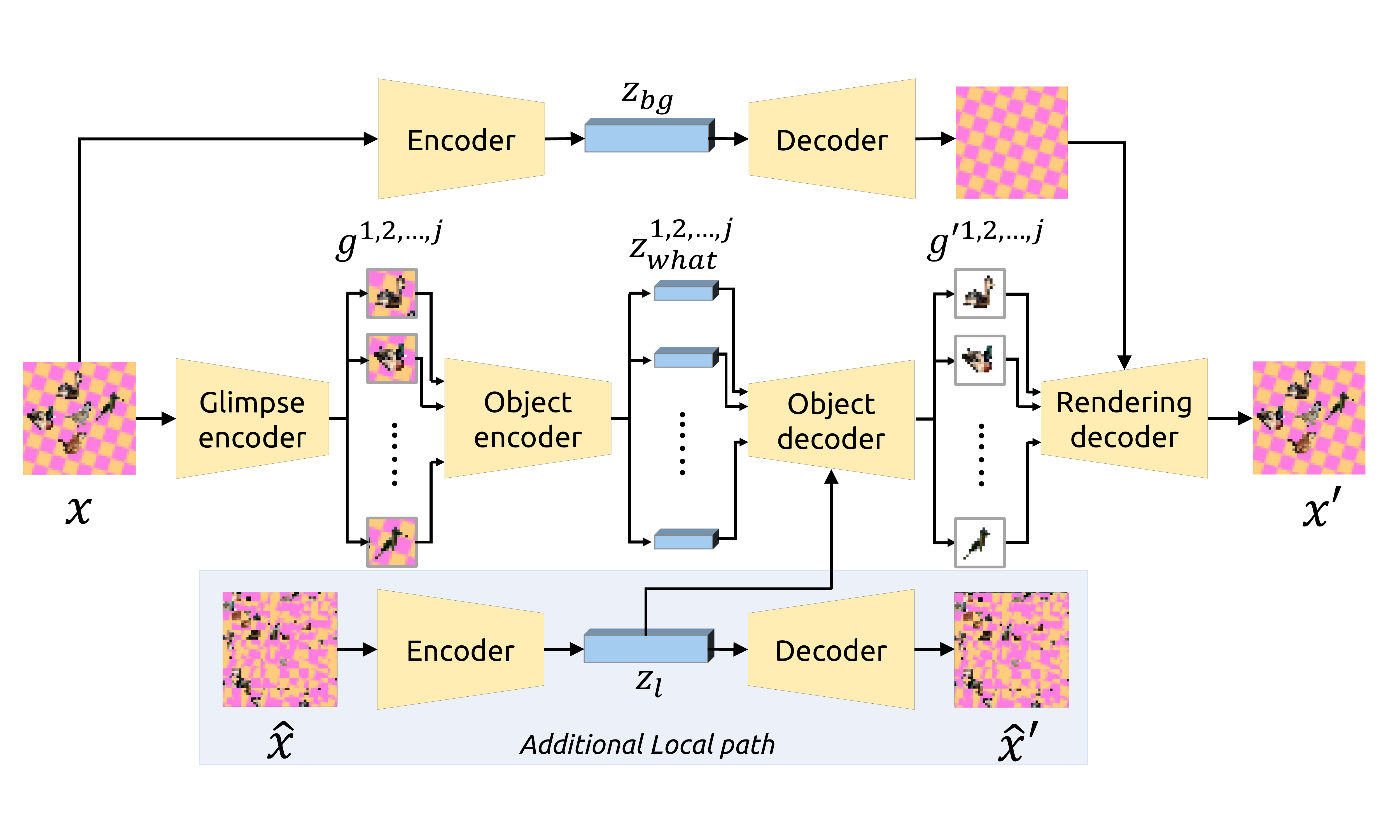}}
    \caption{SPLITED. The SPLIT framework modifies the original VAE architecture only at the decoder's input layer. The additional  autoencoder can be flexibly designed. SPLIT can be applied to the SPAIR model using the same method by adding $z_l$ to the object decoder.
    SPAIR is composed of a glimpse encoder, which selectively looks for multiple objects in the scene using $z_{\text{where}}$. $g^j$ denotes the glimpses, which are patches of pixels that are attended by the glimpse encoder. Next, SPAIR encodes all the glimpses into several $z_{\text{what}}$, which are then reconstructed back into the original image.
    In the figure, other latent variables of the SPAIR model, such as $z_{\text{where}}$ and $z_{\text{pres}}$, are omitted for clarity. SPLIT encourages $z_{\text{where}}$ and $z_{\text{what}}$ to represent object location and identity using only global information. The inductive bias is added in this way because local information such as colour and texture is deemed irrelevant to the object detection task.}
    \label{fig:architecture}
\end{figure}

Our SPLIT framework is a simple extension to the VAE. 
It works by assuming that a subset of latent variables generate auxiliary data in addition to the observed data. The auxiliary data are assumed to contain only local information, which primes the representation of such latent variables to specific information contained in the auxiliary data. The remaining latent variables will avoid representing the same information as they are modelled to be independent factors. In other words, the remaining latent variables are assumed to represent the global context of the data. This is illustrated in Fig. \ref{fig:graphical}. 

Formally, let $x^i$ be the $i^{th}$ observed data point and let $\hat{x}^i$ be the corresponding auxiliary data point that we artificially create (see Sec. \ref{sec:aux_data}), we say that a latent variable $z_{l}^i$ generates $\hat{x}^i$ while both $z_{l}^i$ and $z_{g}^i$ generate the observed data $x^i$.
We assume that $\hat{x}^i$ does not contain global information. This means that $z_{l}$, which dictates the variation in $\hat{x}$, does not need to contain global information in order to generate $\hat{x}$. 

The generation of $x$ is conditioned on both $z_l$ and $z_g$. Since $z_l$ captures the local variation (information contained in $\hat{x}$), $z_g$ has to model the remaining global variation in $x$ that does not exist in $\hat{x}$. $z_g$ and $z_l$ are assumed to be independent. 

This framework can be applied to VAEs with different generative assumptions (Fig. \ref{fig:graphical}). Note that the modification to VAEs is done with a parallel encoder/decoder architecture. Only a minimal modification to the original architecture is required. The resulting model has the flexibility of implicit disentanglement methods while being independent from the optimisation dynamics. This approach opens up possibilities of combining different prior assumptions in a more flexible way. In Sec. \ref{sec:experiments}, we demonstrate the framework by showing how different VAE models can be modified to create the SPLIT representation.

\subsection{Auxiliary Data Transformation}
\label{sec:aux_data}

We are interested in disentangling the global context from local variation in image data. Since $z_l$ is responsible for explaining variation in $\hat{x}$, we would like the auxiliary dataset $\hat{x}$ to only contain local variation of $x$, which would then prime $z_l$ to model the local variation. One way to create $\hat{x}$ with only local information is to destroy the global variation in $x$. 

We propose to destroy the global variation by randomly 
shuffling patches of pixels in the image $x$.
This shuffling transformation is done by first splitting the image 
into patches of $r \times r$ pixels. Each patch is assigned 
an index. The indexes are then randomly 
permuted, and the patches are rearranged accordingly. 
This procedure has two effects: 
(i) local correlations between pixels within each patch 
are preserved, and (ii) global long-range correlations between 
pixels are reduced. 
 
\begin{figure}
    \centering
    \subfloat[\label{a}]{%
       \includegraphics[width=0.15\linewidth]{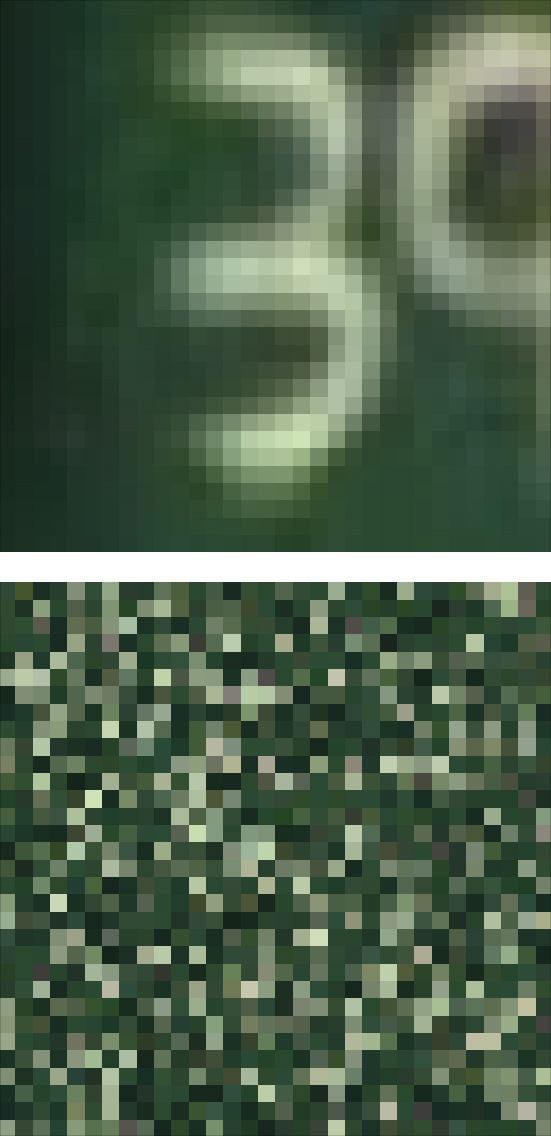}}
       \hspace{1mm}
    \subfloat[\label{b}]{%
        \includegraphics[width=0.15\linewidth]{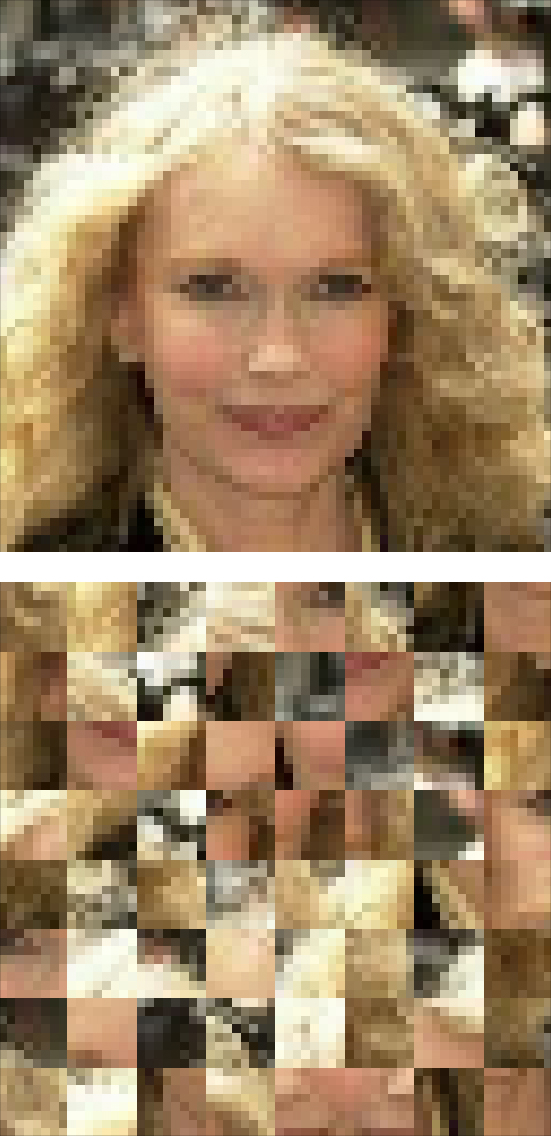}}
        \hspace{1mm}
    \subfloat[\label{c}]{%
        \includegraphics[width=0.15\linewidth]{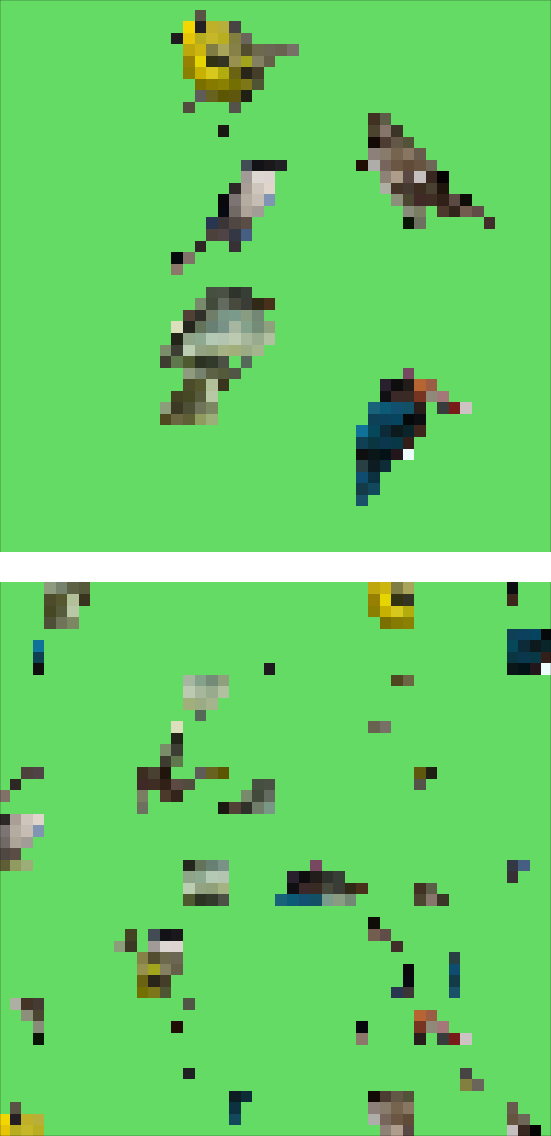}}
        \hspace{1mm}
    \subfloat[\label{d}]{%
        \includegraphics[width=0.15\linewidth]{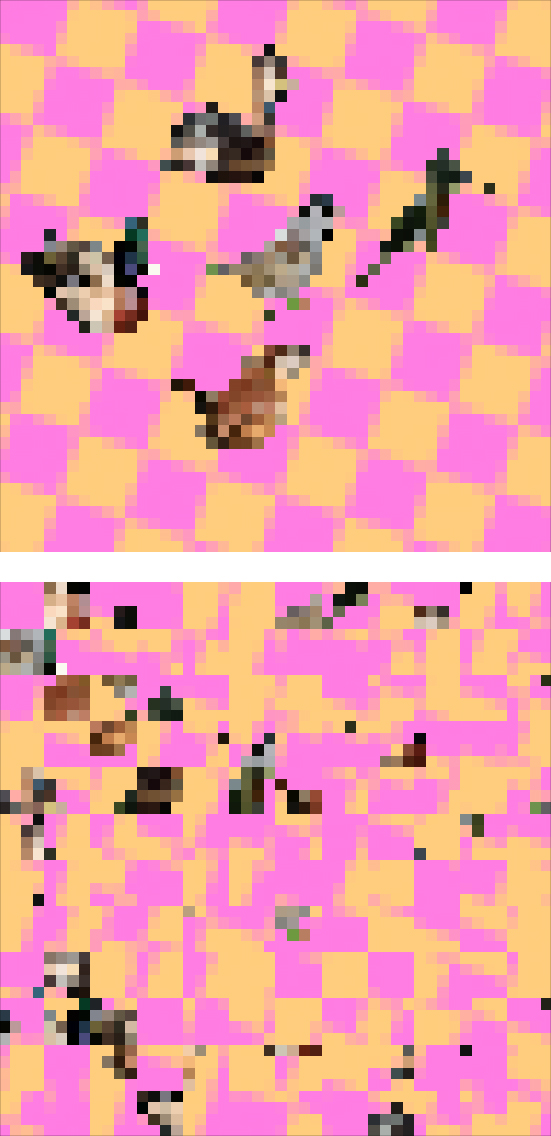}} 
    \caption{Datasets and their auxiliary sets. We use four datasets for our experiments: (a) SVHN, (b) CelebA, (c) Multi-Bird-Easy and (d) Multi-Bird-Hard. The bottom row shows the transformed datasets via the shuffling operation. In the figure, SVHN was shuffled with patch size of one. Multi-Bird datasets were shuffled with patch size of four and CelebA was shuffled with patch size of 8. These patch sizes are chosen for the best disentanglement for each dataset.}
    \label{fig:scramble}
\end{figure}

By transforming the data $x$ in this fashion, the resulting auxiliary dataset $\hat{x}$ 
should not contain any global variation, while it still contains information in the variation of 
local patches, as shown in Fig. \ref{fig:scramble}. 
\section{Experiments}
\label{sec:experiments}

\begin{table*}[t]
\caption{SPLIT-VAE accuracy. We encode and reconstruct the test data with the SPLIT-VAE. Then, we evaluate the reconstructed images with and without modification to the latent space using a pre-trained SVHN classifier. We can observe that changes in $z_g$ result in changes in digit identity, while variations in $z_l$ have only a small effect to the digit identity. The SPLIT-VAE was trained for 300k steps.}
\centering
\scalebox{1.0}{
\begin{tabular}{|c|c|c|c|}
\hline
                                  & \multicolumn{3}{c|}{\textbf{Accuracy(\%)}}   \\ \cline{2-4}
\multicolumn{1}{|c|}{Patch size} &                 &                       &   \\ 
                                  & Direct reconstruction          & Reconstruction with $z_l \sim \mathcal{N}(0,1)$    & Reconstruction with $z_g \sim \mathcal{N}(0,1)$     \\ \hline
                                  &                         &                                     &               \\
1                    & $98.39$ $(\pm 0.05)$  & $87.28$ $(\pm 0.34)$                  & $13.26$ $(\pm 0.73)$                  \\
2                    & $98.37$ $(\pm 0.04)$  & $84.43$ $(\pm 1.44)$                  & $11.23$ $(\pm 0.43)$                  \\
4                    & $98.42$ $(\pm 0.05)$  & $73.56$ $(\pm 2.55)$                  & $10.82$ $(\pm 0.35)$                  \\
8                    & $98.56$ $(\pm 0.04)$  & $59.19$ $(\pm 2.43)$                  & $11.03$ $(\pm 0.36)$                 \\ \hline   
\end{tabular}}

\label{table:vae_acc}
\end{table*}


The aims of our experiments\footnote{Each experiment is repeated 10 times. We report the results with one standard deviation, denoted by $\pm$.} are as follows.
\begin{itemize}
    \item To verify that the proposed SPLIT framework can be applied to a variety of VAE models with different latent structures.
    \item To evaluate whether the resulting models can effectively perform disentanglement of local and global information.
    \item To demonstrate that the SPLIT framework can improve interpretability of the learnt representations.
    \item To demonstrate that the disentangled representations can be use to help improve generalisation and robustness in certain downstream tasks by imposing useful inductive biases into the model.
\end{itemize}


\subsection{Datasets}
We use the following datasets: 
\begin{itemize}
    \item SVHN : SVHN \cite{netzer2011reading} consists of 32x32 RGB images of street numbers from Google Street View.
    \item CelebA : We use a cropped version of face images from CelebA dataset \cite{liu2015faceattributes}. CelebA is a face dataset with a variety of pose and background variations. We further crop and resize the images into 64x64 RGB images.
    \item Multi-Bird-Easy : A dataset is created from bird instances extracted from the CUB dataset \cite{WahCUB_200_2011} and resized into 14x14 RGB images. The bird instances are then randomly placed onto 48x48 plain-colour RGB images. Each image contained 0-5 birds. The training set consists of 8 background colours, while the test set has four unseen background colours.
    \item Multi-Birds-Hard : Identical to Multi-Bird-Easy except that the backgrounds are randomly rotated checkerboards of two colours instead of a plain background. There are six background colours in the training set and three unseen background colours in the test set.
\end{itemize}

\subsection{Local-global disentanglement in vanilla VAE}

\begin{figure*}[!t]
    \centering
    \subfloat[Vary both $z_g$ and $z_l$\label{2a}]{%
       \includegraphics[width=0.25\linewidth]{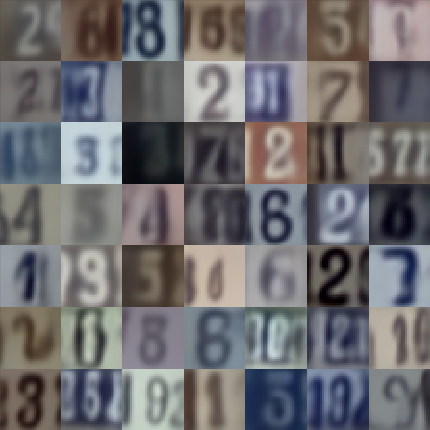}}
       \hspace{0.5cm}
    \subfloat[Vary $z_l$\label{2b}]{%
        \includegraphics[width=0.25\linewidth]{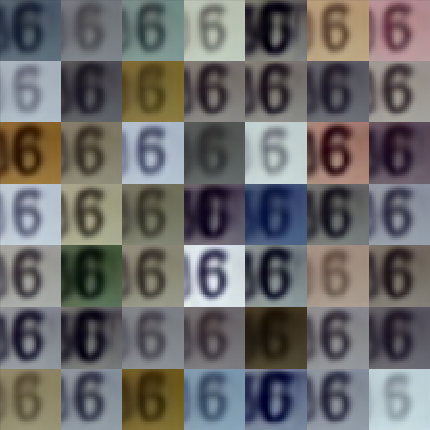}}
       \hspace{0.5cm}
    \subfloat[Vary $z_g$\label{2c}]{%
        \includegraphics[width=0.25\linewidth]{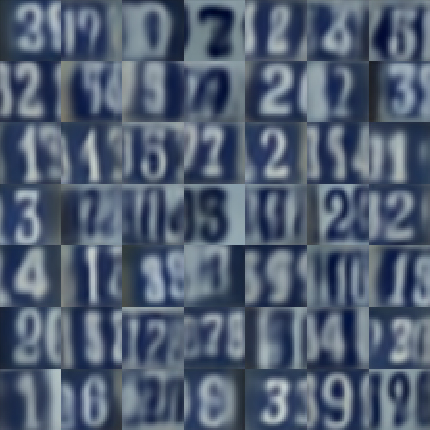}}
   \\
   \subfloat[Vary both $z_g$ and $z_l$\label{2d}]{%
       \includegraphics[width=0.25\linewidth]{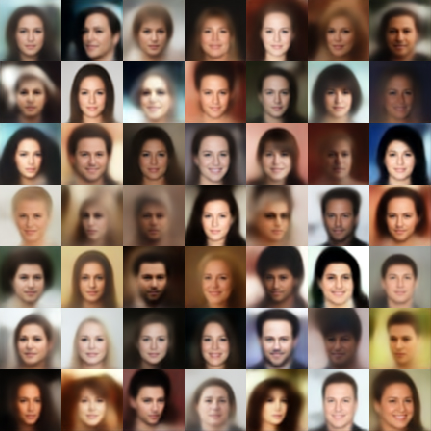}}
       \hspace{0.5cm}
    \subfloat[Vary $z_l$ \label{2e}]{%
        \includegraphics[width=0.25\linewidth]{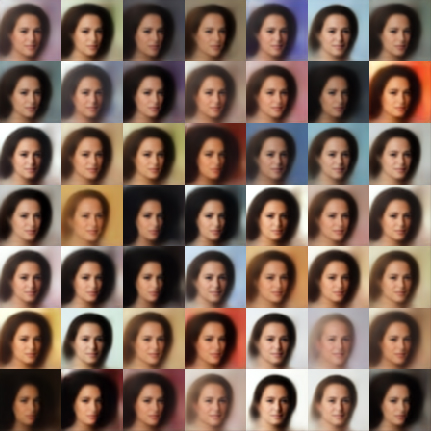}}
        \hspace{0.5cm}
    \subfloat[Vary $z_g$ \label{2f}]{%
        \includegraphics[width=0.25\linewidth]{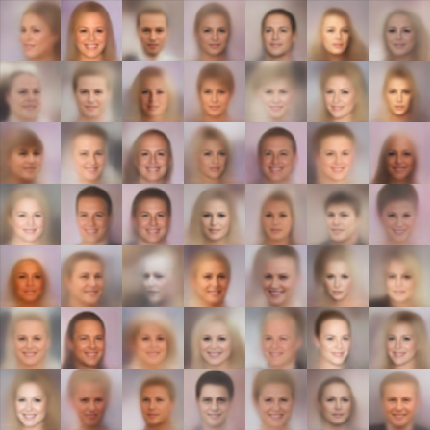}}
    \caption{Visual inspection of the generated samples from SPLIT-VAE. We visualise the meaning of the representations $z_l$ and $z_g$ via generated samples. As expected, variations in $z_g$ correspond to the change in images' global structure such as digit identity and global style for SVHN and face orientation and gender for CelebA, while variations in $z_l$ dictate the colour patterns of the images.}
    \label{fig:generate_split_vae}
\end{figure*}

We first investigate the local-global disentanglement in a vanilla VAE. We use the standard VAE generative assumption which can be combined with an auxiliary data generation process as follow:
\begin{align}
    z_l, z_g &\sim p(z_l, z_g), \\
    x &\sim p_{\theta}(x| z_l, z_g), \\
    \hat{x} &\sim p_{\hat{\theta}}(\hat{x}| z_l).
\end{align}

We call this model SPLIT-VAE. We use encoder $q_{\hat{\phi}}(z_l | \hat{x})$ and decoder networks $p_{\hat{\theta}}(\hat{x}| z_l)$ for the local variables in addition to the standard VAE encoder $q_{\phi}(z_g | x)$  and decoder $p_{\theta}(x|z_g, z_l)$.

We optimise $\phi$, $\hat{\phi}$, $\theta$ and $\hat{\theta}$ using the standard Monte Carlo estimate of the ELBO,
\begin{align}
 \mathcal{L} &= \log p_{\theta}(x| z_l, z_g) + \log p_{\hat{\theta}}(\hat{x}| z_l)\nonumber \\
    &- \beta KL\left( q_{\phi, \hat{\phi}}(z_g, z_l|x ,\hat{x}) || p(z_l, z_g)\right),
\end{align}
where $\beta>0$ is a hyper-parameter for adjusting the compression terms, which have been shown to help improve disentanglement~\cite{higgins2016beta}. The prior $p(z_l,z_g)$ is a unit diagonal Gaussian.

Fig. \ref{fig:architecture} illustrates the neural network architectures of the original VAE and the SPLIT-VAE. The original VAE composes of an encoder and a decoder. The SPLIT-VAE has an extra encoder-decoder pair while leaving the architecture of the original VAE almost untouched except for the input size of the decoder. The extra encoder is the variational posterior $q_{\hat{\phi}}(z_l|\hat{x})$ and the extra decoder is the likelihood $p_{\hat{\theta}}(\hat{x}|z_l)$.

\subsubsection{Visual inspection of the generative samples}
One way to assess the quality of the learnt latent variables is to inspect samples generated from the model. We would like to see how the generated data vary with different values of $z_l$ and $z_g$. 
From the samples generated from models trained with the SVHN or CelebA dataset (Fig. \ref{fig:generate_split_vae}), we see that varying $z_l$ generally gives the data variations in colour pattern of the background, digits and faces. On the other hand, varying $z_g$ produces images at semantic levels, such as the digit identity, face orientation or gender.  



\subsubsection{Quantitative inspection using a trained classifier}
In order to show more \emph{quantitatively} that the method can explicitly place global information in a subset of the latent variables, an experiment is carried out where an SVHN classifier is trained to inspect the generative samples of the model. The classifier has an accuracy of 99.66\% on the SVHN test set by training on both the test and training set. The SPLIT-VAE, however, only sees the training set.

The SVHN test data are encoded into the latent space with the encoder of the model used in the previous experiment with $\beta=1$. 
Then, three types of samples are generated from the encoding: (i) images generated directly from the encoded latents, (ii) images in which $z_l$ is replaced with a random sample from $\mathcal{N}(0,1)$ while preserving $z_g$, and (iii) images in which $z_g$ is replaced with a random sample from $\mathcal{N}(0,1)$ while preserving $z_l$.

The result in Table \ref{table:vae_acc} shows that: (i) the direct reconstruction slightly perturbs digit identity, yielding an accuracy of $98\%$, (ii) varying $z_l$ also slightly perturbs the digit identity, yielding $87\%$ accuracy, while (iii) varying $z_g$ completely changes the identity of the digits, reducing the accuracy to $13\%$. The difference between $87\%$ and $13\%$ (i.e., chance) in (ii) and (iii) demonstrates quantitatively the disentangling we were aiming for.

We also investigate the sensitivity to the shuffle patch size ($r$). We find that a patch size from one to four can disentangle SVHN digit from its colour effectively. The larger patch sizes can result in $z_l$ occasionally being used to represent the digit identity. This result can be interpreted as the framework was able to only disentangle the occurrence frequency of each colour in an image. Interestingly for CelebA, by visual inspection, we find that a patch size too small can worsen the disentanglement. In Fig. \ref{fig:generate_split_vae}, we use a patch size of 8 to generate sample for CelebA, which is tuned via inspection to create the best looking disentanglement. With a larger patch size, we observe that $z_l$ can also represent the local correlations such as the simple local colour gradient and colour pattern in nearby pixels. 

\subsubsection{Experiment investigates the representation for style-transfer task}

\begin{figure}
    \centering
    \subfloat[\label{3a} SVHN]{%
       \includegraphics[width=0.8\linewidth]{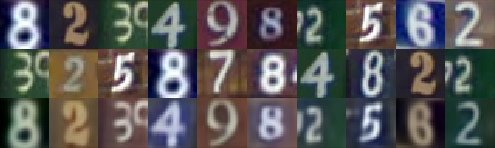}} \hspace{1cm}
    \\
    \subfloat[\label{3b} CelebA]{%
       \includegraphics[width=0.8\linewidth]{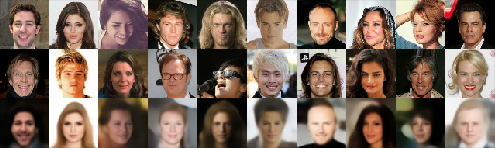}}
    \caption{Style transfer using SPLIT-VAE. We take $z_g$ from the top row and $z_l$ from the middle row to generate images in the bottom row. For SVHN, we can see the style corresponding to the colour pattern (a). For CelebA, we can see that the style corresponding to the hair, skin tone and background colour (b).}
    \label{fig:ex_style}
\end{figure}

In this experiment, we investigate the learnt local representation in the task of style transfer. Similar to the previous experiments, we encode the data using the SPLIT-VAE and change their encoded $z_l$ to see the resulting generated data. we transfer $z_l$ encoded in one image to another. From visual inspection of the result, Fig. \ref{fig:ex_style}, the local style of one image is successfully transferred to the other. This further confirms that $z_l$ represented the local information.

\subsection{Clustering the global information using Gaussian-Mixture VAE}
\begin{figure*}
    \centering
    \subfloat[Vary $z_g$ from the same cluster\label{4a}]{%
        \includegraphics[width=0.24\linewidth]{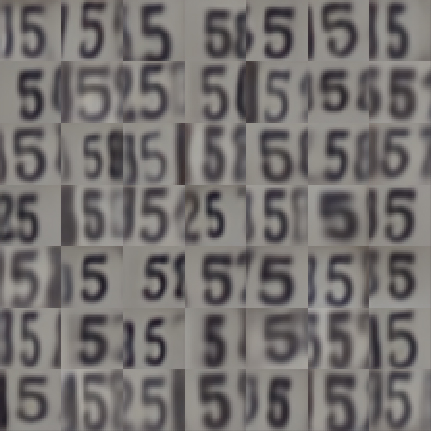}} \hfill
    \subfloat[Each row are generated from the same cluster $y$\label{4b}]{%
        \includegraphics[width=0.24\linewidth]{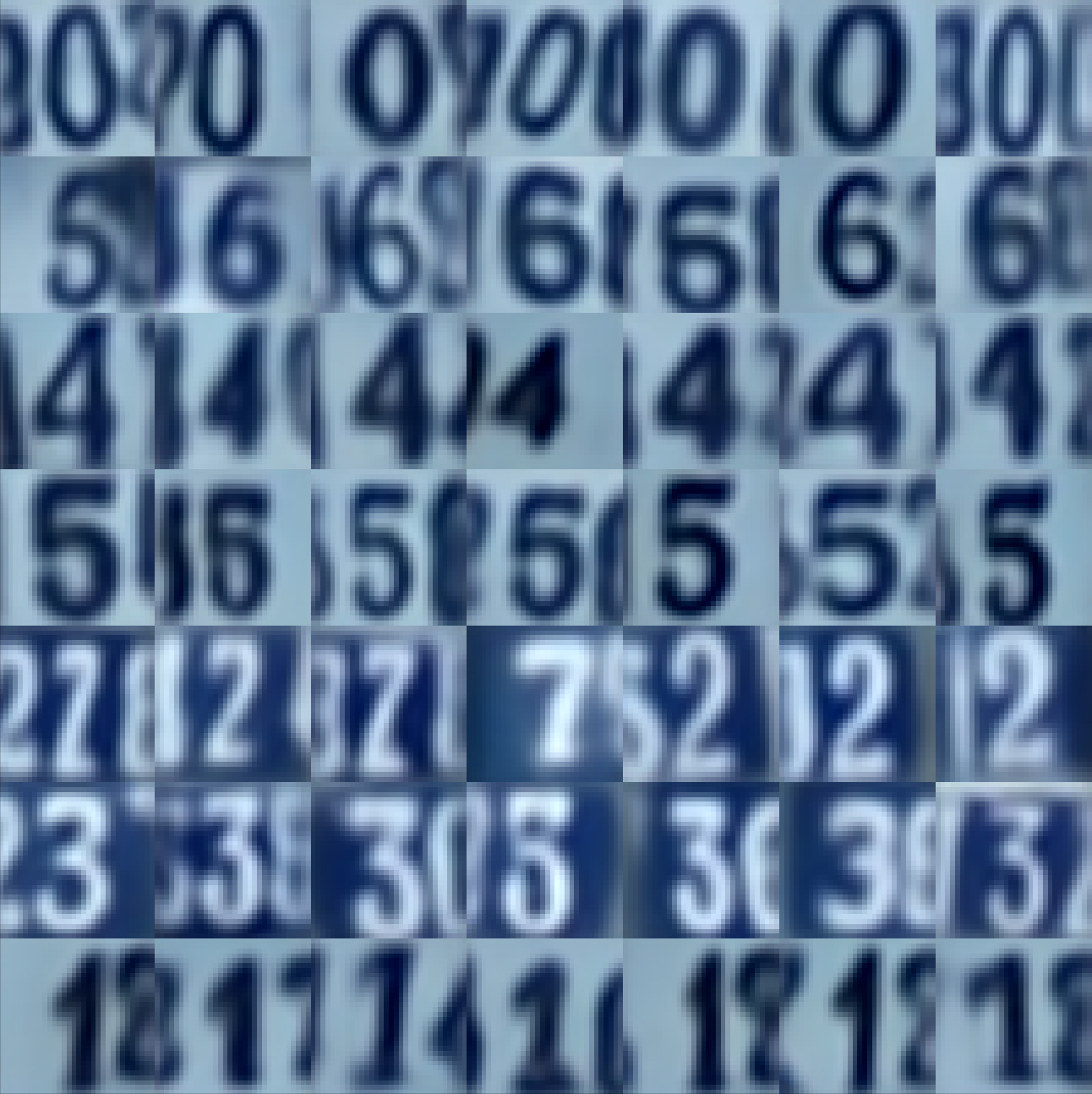}} \hfill
    \subfloat[Vary $z_g$ from the same cluster\label{4c}]{%
        \includegraphics[width=0.24\linewidth]{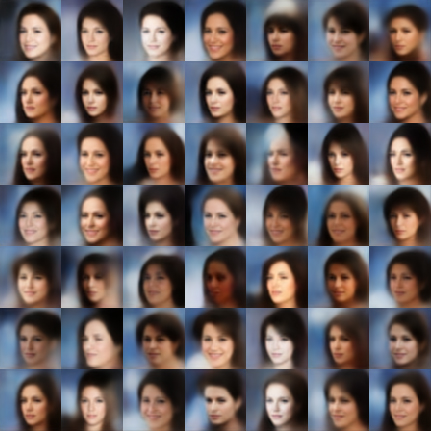}} \hfill
    \subfloat[Each row are generated from the same cluster $y$ \label{4d}]{%
        \includegraphics[width=0.24\linewidth]{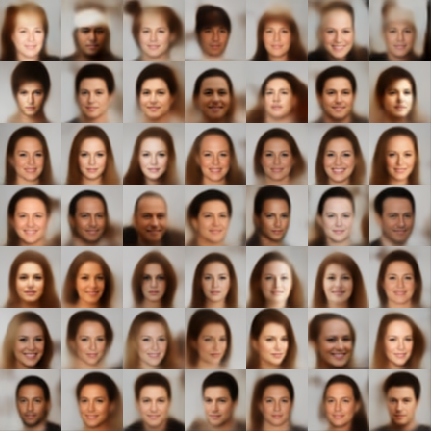}}
    
    \caption{Visualising the representation of SPLIT-GMVAE. Similar to the previous experiment, we visualise the meaning of $z_g$ and $y$ by inspecting the generated samples at different values of $z_g$ and $y$, while keeping $z_l$ the same. For SVHN, we can see that $y$ represents digit identity, while $z_g$ represents the digit style and orientation. For CelebA, $z_g$ controls some facial features, while $y$ controls gender and face orientation. $z_l$ controls the background colour.}
    \label{fig:visualise-split-gmvae}
\end{figure*}

In this experiment, we explore a simple extension to the VAE, namely Gaussian-Mixture VAE (GMVAE). As described in Sec. \ref{sec:gmvae}, GMVAE can represent data with both a continuous latent variable $z$ and a discrete latent variable $y$, allowing it to do clustering by encoding data using $y$. By applying the SPLIT framework to the GMVAE, we assume the following generative process, where we impose an inductive bias on the cluster variable $y$ to only affect the global information $z_g$:
\begin{align}
z_l      &\sim p(z_l), \\
z_g, y   &\sim p_{\gamma}(z_g |y)p(y), \\
x 	     &\sim p_{\theta}(x| z_g, z_l), \\
\hat{x}  &\sim p_{\hat{\theta}}(\hat{x}| z_l).
\end{align}

We assume the variational posterior factorised as $q_{\phi_g}(y, z_g |x )q_{\phi_l}(z_l |\hat{x})$. We use diagonal Gaussians as the posteriors of continuous variables $z_g$, $z_l$ and a Gumbel-Softmax (Concrete) distribution ~\cite{maddison2016concrete,jang2016categorical} for the class variable $y$ with a constant temperature $\tau$. Similar to SPLIT-VAE, we optimise the following ELBO objective:
\begin{align}
\mathcal{L} &= \log p_{\theta}(x| z_l, z_g) + \log p_{\hat{\theta}}(\hat{x}| z_l)   \nonumber \\
    &- \alpha KL\left(q_{\phi_g}(y|x)||p(y)\right) \nonumber \\
    &- \beta KL\left( q_{\phi_g}(z_g|x ,y)q_{\hat{\phi_l}}(z_l|\hat{x}) || p_{\gamma}(z_g|y)p({z_l})\right),
\end{align}
where $\alpha,\beta>0$ are hyper-parameters controlling the level of influence by the prior terms.

\subsubsection{Visual Inspection of the latent representation}

We investigate the latent space by visualising the generated samples from different latent values. Fig. \ref{fig:visualise-split-gmvae} shows that, for SVHN, the clusters represent digit identity. In contrast to previous work \cite{dilokthanakul2016deep}, deep clustering of SVHN is shown to result in clusters with similar background colours rather than the same digit class. SPLIT-GMVAE is biased towards clusters of global information, which helps improve the interpretability of the clusters as information of interest is generally separated in different scales. For CelebA, the biased clusters represent gender information and face orientation. 

\subsubsection{Clustering of unseen data}

\begin{figure}
    \centering
    \subfloat[SVHN]{
        \includegraphics[width=0.4\linewidth]{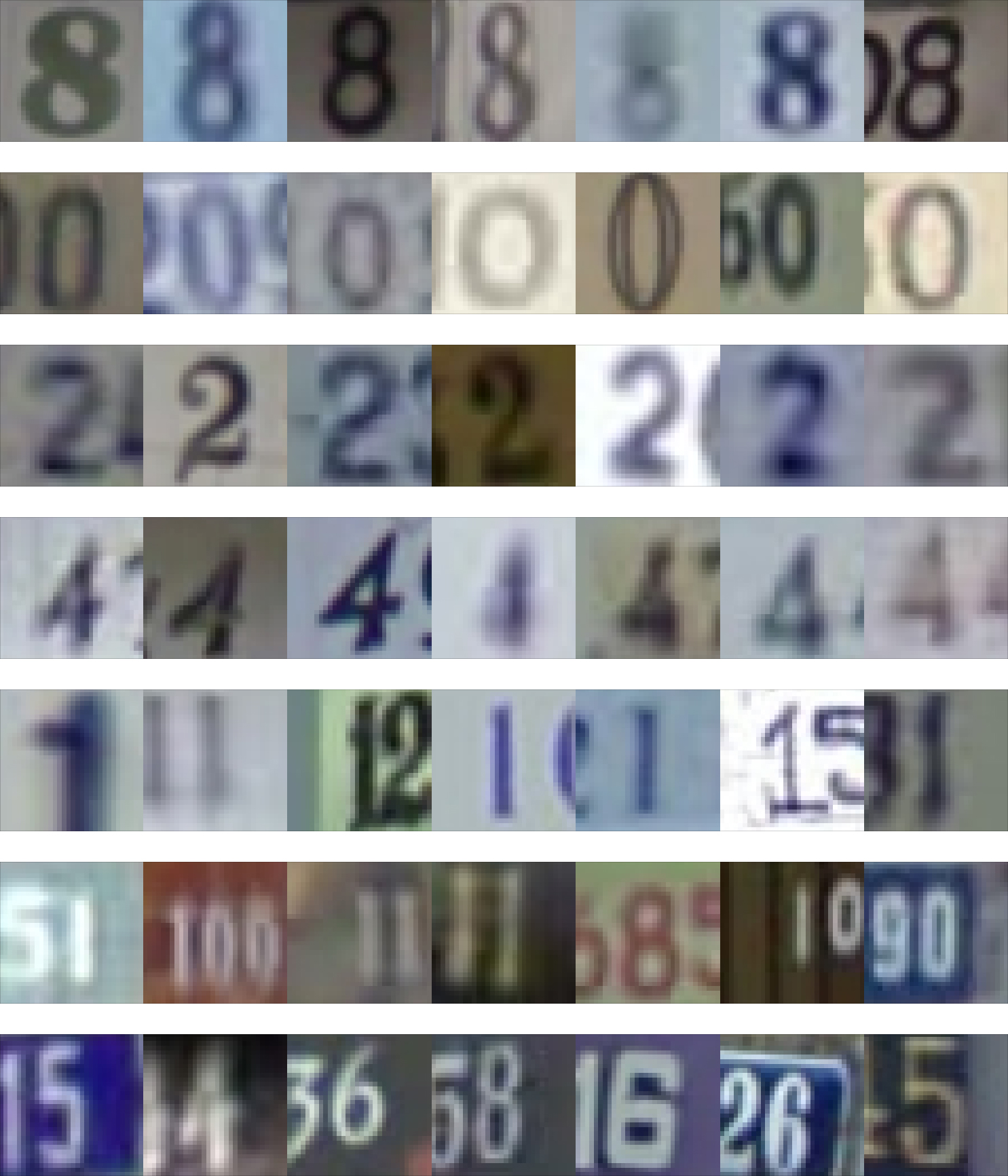}
    } 
    \subfloat[CelebA]{
        \includegraphics[width=0.4\linewidth]{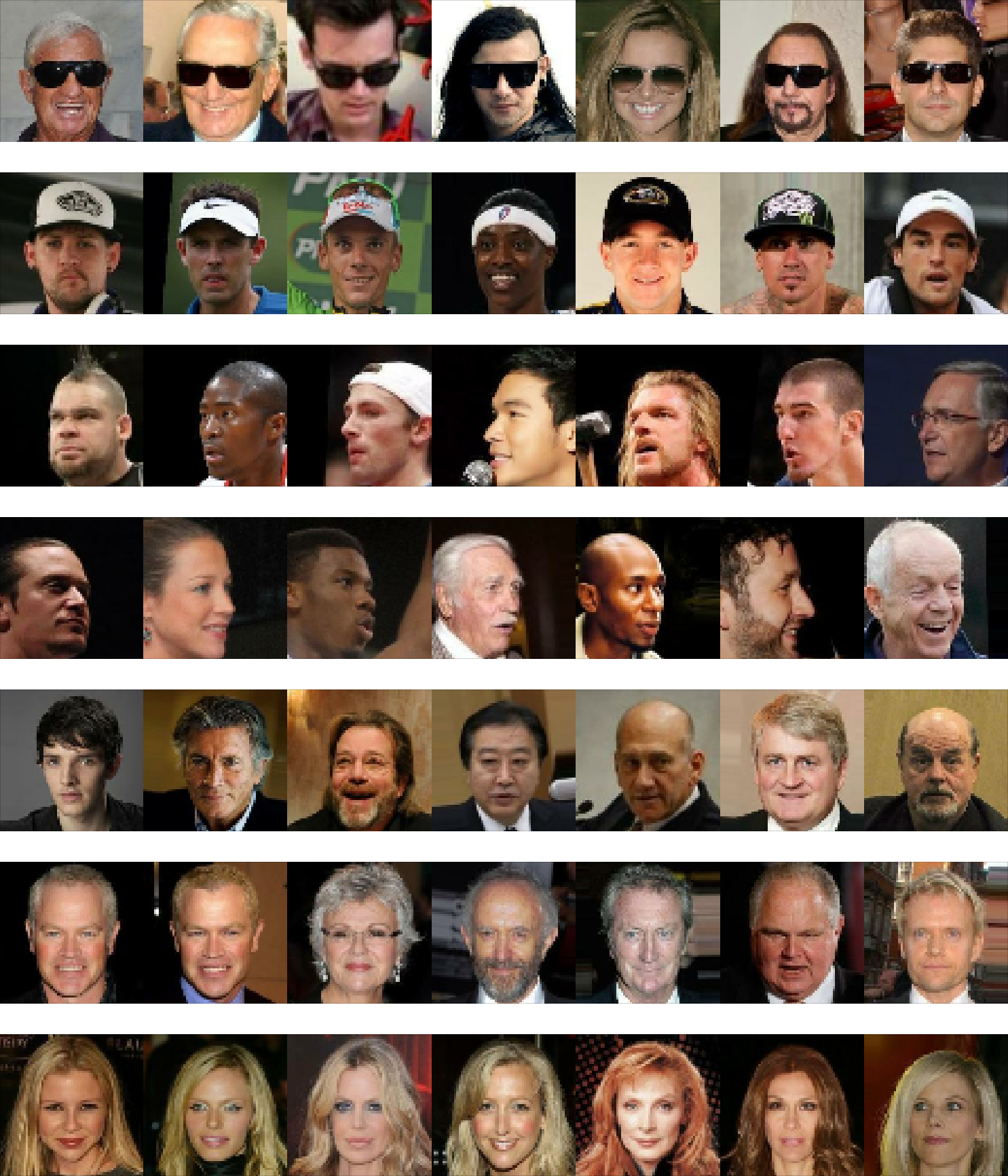}
    }
    \caption{Visualisation of clustering results from unseen data by SPLIT-GMVAE. Each row contains samples assigned to the same cluster. Results from SVHN dataset show that each cluster represented global variation in the images such as the digit identity or the presence of side digits (a). Similar results were shown for CelebA dataset. A cluster may represent the presence of accessories (e.g. sunglasses or hat), face orientation, gender or hair style (b). }
    \label{fig:unseen_clustering}
\end{figure}

We have visualised the generated data in the previous experiments, which shows the meaning of the values of each latent variable. In this experiment, we investigate the ability to infer such values from unseen data by the encoder. We encode the test data using the trained SPLIT-GMVAE and then inspect the resulting cluster encoded. SPLIT-GMVAE display its capability to cluster unseen data with similar global structure together in both SVHN and CelebA dataset as shown in Fig. \ref{fig:unseen_clustering}. We can see that the results are based on global variation of each dataset while local variations (e.g. colour) are mostly ignored. Hence the data assigned to each cluster could have a variety of colours and styles but are similar in terms of global structure.

\subsubsection{Clustering Accuracy of SPLIT-GMVAE}

Next, we evaluate the SPLIT-GMVAE with SVHN unsupervised clustering benchmark. We measure the clustering accuracy (ACC) \cite{hu2017learning} on the test set of SVHN. The ACC works by assigning the label to each cluster according to the majority of the true class of samples inside the cluster, then simply measure the percentage of the correctly assigned class. A high ACC score means that, for each cluster represented by the model, the classes of digit images inside the cluster are the same. 

Importantly, we believe this benchmark is not useful comparing clustering algorithms. Because a good clustering algorithm can also be good at clustering local features such as background colour, where this score deems irrelevant. However, this benchmark is a moderately good benchmark for investigating the inductive bias in the clustering algorithm. For SPLIT-GMVAE, the model is biased towards clusters of global information. Since the digit identity is arguably the most dominant global information in the SVHN dataset, higher scores would mean better disentanglement by the SPLIT-GMVAE\footnote{SPLIT-GMVAE was searched for the best parameters using $\alpha$=\{1, 5, 10, 20, 30, 40, 50, 60\} and $\beta$=\{1, 10, 20, 30, 40, 50, 60, 70, 80\}, $\tau$=0.1-0.8\ and patch size=\{1, 4, 8\}. The best setting is $\alpha=40$, $\beta=40$, $\tau=0.4$ and patch size=4. Performance gets better as $k$ increases.}. We report the clustering result in Table. \ref{table:cluster_acc} showing a strong inductive bias in the SPLIT-GMVAE, resulting in better clustering of SVHN digits compared with the GMVAE. We note that there are other global information that does not describe the digit identity, i.e. the presence of side digits. This reduces the ACC score (see Fig. \ref{fig:unseen_clustering}).

\begin{table}
\caption{SVNH clustering ACC score. We trained GMVAE and SPLIT-GMVAE for 3M steps and evaluated on the SVHN test data using the clustering ACC score. The results for DEC, IMSAT and ACOL-GAR are taken directly from their papers. K is the number of clusters.
}
\centering
\begin{tabular}{l|c|c}
\multicolumn{1}{c|}{\textbf{Model}}     & \multicolumn{1}{c|}{\textbf{K}} & \multicolumn{1}{c}{\textbf{ACC(\%)}}  \\ \hline
                   &            &  \\
DEC~\cite{xie2016unsupervised}               & 10         & $11.9$ $(\pm 0.4)$  \\
IMSAT~\cite{hu2017learning}               & 10         & $57.3$ $(\pm 3.9)$  \\
ACOL-GAR~\cite{kilinc2018learning}   & 10         & $76.8$ $(\pm 1.3)$  \\
\hline
GMVAE~\cite{dilokthanakul2016deep,shu2017note}              & 10         & $35.7$ $(\pm 9.9)$  \\
\textbf{SPLIT-GMVAE}     & 10         & $40.8$ $(\pm 10.9)$  \\
\hline
GMVAE              & 30         & $61.1$ $(\pm 3.9)$  \\
\textbf{SPLIT-GMVAE}     & 30         & $73.5$ $(\pm 4.6)$  \\
\end{tabular}

\label{table:cluster_acc}
\end{table}

\subsection{Effects of local-global disentanglement in unsupervised object detection model}

Another experiment is done to examine the effects of local-global disentanglement in the unsupervised object detection model SPAIR. This experiment is done using two different datasets, Multi-Bird-Easy and Multi-Bird-Hard.
Since SPAIR has an explicit latent variable for object existence $z_{\text{pres}}$, we can use this latent representation to evaluate object counting accuracy directly. 

We implement SPAIR \cite{crawford2019spatially} with an additional background model. Our background model is an encoder-decoder model, which encodes an image into $z_{\text{bg}}$ and uses a decoder to create a background image. The reconstructed background is merged with the reconstructed canvas of objects, resulting in a full reconstruction. This is then used to compute the usual ELBO loss.  
SPLIT-SPAIR is implemented the same way as SPAIR \cite{crawford2019spatially} but with an additional encoder-decoder pair (Fig. \ref{fig:architecture}). Similar to other SPLIT models, the encoder takes $\hat{x}$ as input and encodes it into $z_l$, which is then reconstructed back into $\hat{x}$. Here, $z_l$ is concatenated with each of the $z_{\text{what}}$ and is used for reconstruction of the object glimpses.

\subsubsection{Counting Performance of SPAIR and SPLIT-SPAIR}

\begin{figure}
    \centering
    \includegraphics[width=0.8\linewidth]{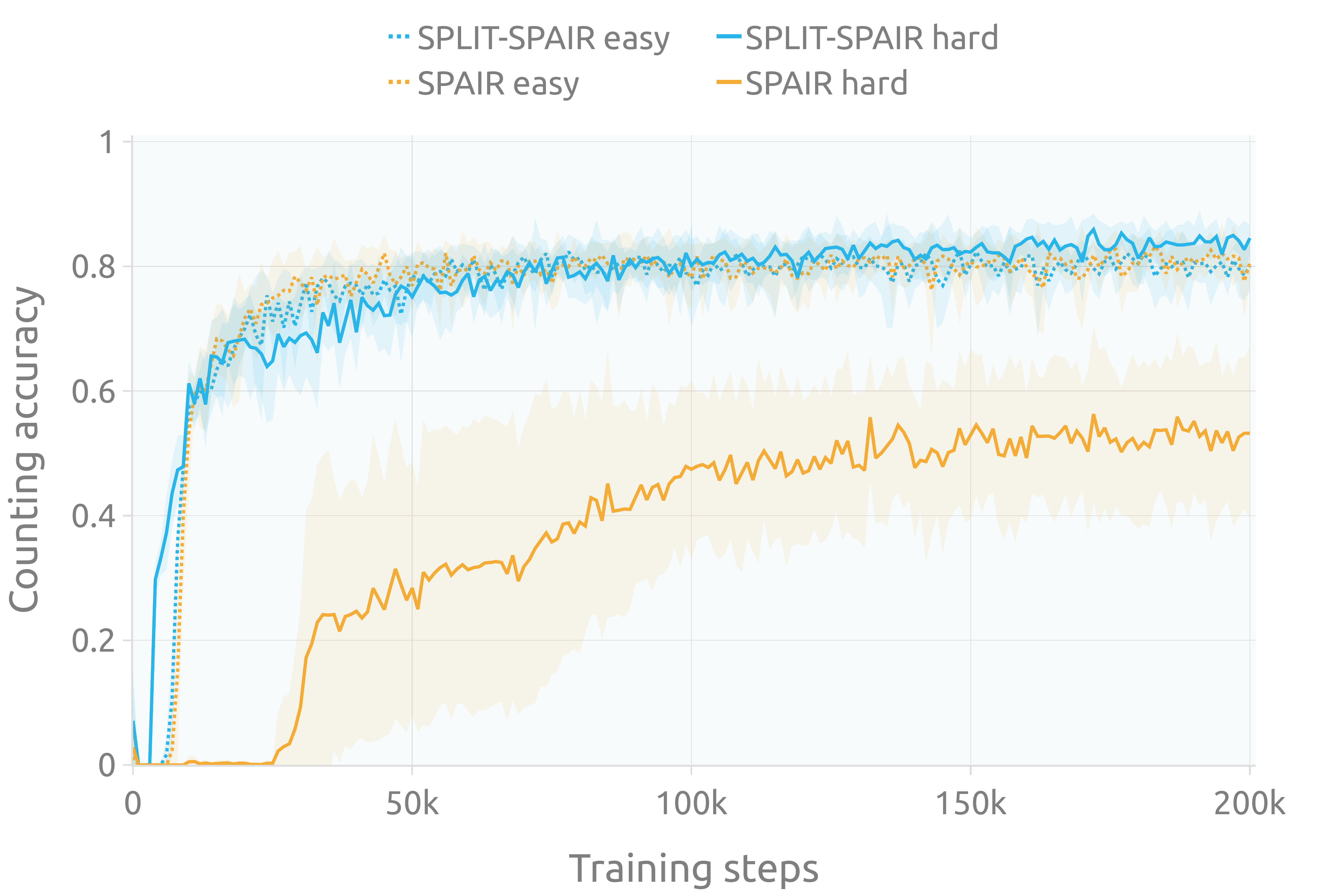}
    \caption{Counting accuracy of SPAIR and SPLIT-SPAIR in the test set of Multi-Bird-Easy and Multi-Bird-Hard datasets. Both models achieved $>$80\% accuracy in the easier dataset, but SPLIT-SPAIR substantially outperformed SPAIR on Multi-Bird-Hard with harder background images.}
    \label{fig:spair_acc}
\end{figure}

One simple benchmark of object detection performance is the counting accuracy. The goal of this task is to count the number of objects in a given image. To do this, the model needs to differentiate foreground objects from the background image. Both SPAIR and SPLIT-SPAIR exhibit strong performance when trained and tested in Multi-Bird-Easy dataset, exceeding 80\% counting accuracy. The effects of the SPLIT framework are shown when we compare them on the Multi-Bird-Hard dataset. There the background images become more complex, making it harder for the models to distinguish objects from background. SPLIT-SPAIR maintains the same level of performance, but this is not the case for SPAIR when trained and tested on the Multi-Bird-Hard dataset as shown in Fig. \ref{fig:spair_acc}. This confirms that disentangled representations are certainly useful for downstream tasks especially when the data is more complex.

\begin{figure}
    \centering
    \includegraphics[width=1.0\linewidth]{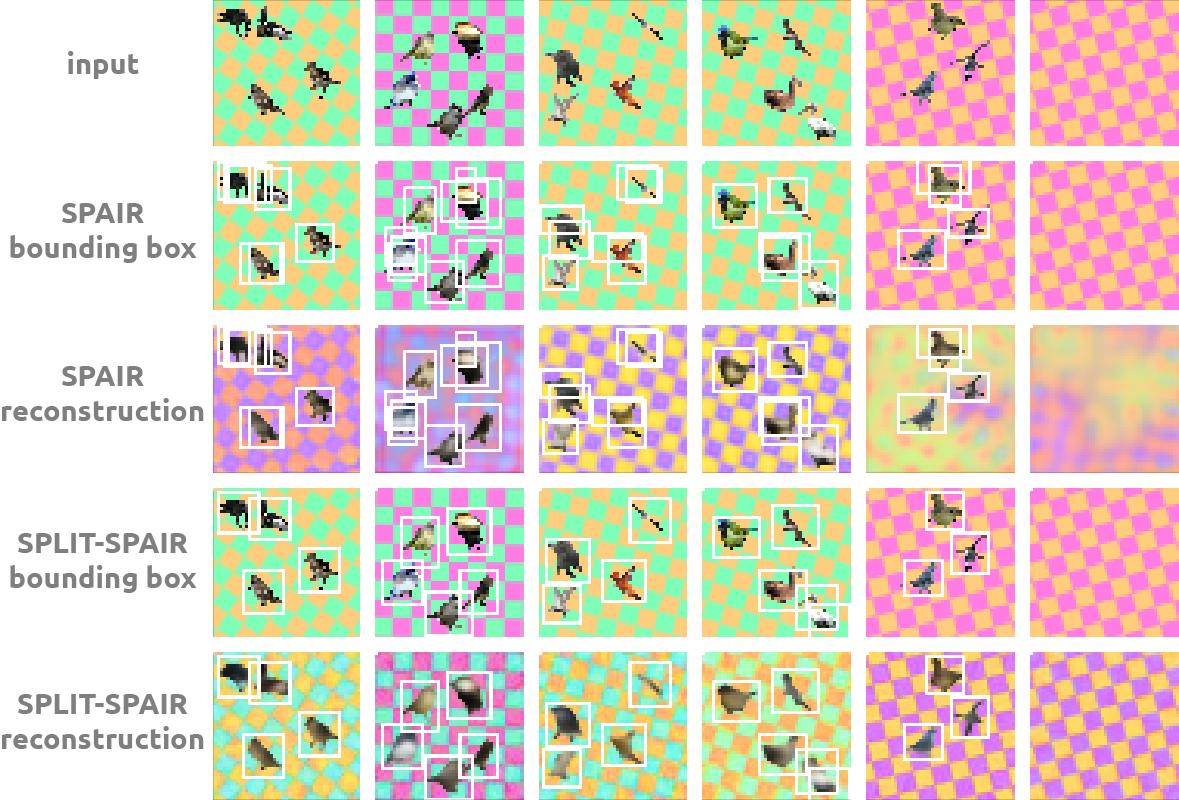}
    \caption{Examples of predicted bounding boxes  and reconstruction of SPAIR and SPLIT-SPAIR on the Multi-Bird-Hard dataset. SPAIR produced multiple bounding boxes on the same object while SPLIT-SPAIR was consistently more accurate. Similarly for the background reconstruction task SPAIR was unable to generalise to unseen data (background colour) and produced colours that did not match the input colours, while SPLIT-SPAIR output colours that were more similar to the inputs.}
    \label{fig:bbox_compare}
\end{figure}

We further investigate why and how SPLIT-SPAIR is more robust than SPAIR. From Fig. \ref{fig:bbox_compare}, we can see that SPAIR is distracted by unseen background colours and struggle to output the correct colours. SPLIT-SPAIR, however, manage to generalise from training data and produce colours that are closer to the inputs than SPAIR. The results illustrate that explicitly imposing this granularity bias in SPAIR gives the model an ability to ignore variation in the background in the object representation path. This yields a more robust object detection model, which in turn helps the background model to better learn to infer background features.

\section{Result Discussion}

The results show successful disentanglement of local and global information by the SPLIT framework in the three VAE models. The framework is generic, allowing imposition of inductive biases in arbitrarily complex models such as SPAIR improving both learnt representation and learning ability. However, there are several limitations that warrant further investigation. 

First, the framework introduces an additional autoencoder network. This doubles the memory cost and requires a significant amount of additional computing power for both forward and backward passes. An interesting open question is whether there are other objectives that can achieve a similar local-global disentanglement effect without the need need to train a large auxiliary model.

Second, the local variation learnt by the SPLIT framework seems to be limited to colour patterns. Although disentanglement of such patterns can unveil much more interpretable features in the global representation, the local variation in $z_l$ seems to be largely limited. It will be interesting to investigate further on why this is the case, and how to encode more complex local information such as texture into $z_l$ under this framework. 

Third, the effect of the patch size parameter on disentanglement remains to be understood. We found that a very small patch size, e.g. one or two, worked well for the SVHN but not for the CelebA or the Multi-Bird datasets. However, a patch size that is too big can result in worse disentanglement.   

Forth, we used a simple data transformation, namely shuffling of image patches, to remove global long-range structural correlations in the input images. Other transformations could easily be conceived, for example, image blurring, texture scrambling \cite{PortSimo2000} and phase scrambling \cite{WatsHartAndr2014}. It is also possible achieve a more fine-grained disentanglement by using additional data transformations that preserve different levels of correlations, such as scrambling within image patches \cite{WatsHartAndr2017}. Moreover, it remains to be explored as to what kind of information is lost and what is preserved, e.g. in terms of natural image statistics or other visual features, by different transformations. We leave this for future work.

Finally, the SPLIT-SPAIR is found to improve the SPAIR's object detection ability to generalise to unseen background colours as well as the ability to interpolate checkerboard orientations. However, it failed extrapolate to unseen background textures. We believe that better understanding of the out-of-distribution extrapolation performance is one of the most important research directions that could improve the success of future machine learning algorithms.    

\section{Disentanglement in Cognitive Neuroscience}
The compositional representations of local or low-level features and global or higher-level  features in deep generative models can be seen to have parallel neural representations in the visual systems of humans and non-human primates \cite{CichKais2019, EpstBake2019, CichKhosPant+2016, RichLillBeau+2019, DiCaCox2007, KarKubiSchm+2019}. To support higher-level visual processing like object, face, and scene perception, the visual system has to be able to disentangle neural codes that underlie the representations of these low- and higher-level features. 

Evidence from cognitive neuroscience research has demonstrated that the ventral visual pathway plays many critical roles in untangling information about an object or a scene \cite{EpstBake2019, DiCaZoccRust2012, DiCaCox2007}. The ventral visual pathway comprises of multiple visual areas that are interconnected in a hierarchical fashion \cite{FellEsse1991, DiCaCox2007}. The up-stream visual areas including primary and extrastriate visual cortices (i.e., V1, V2, V3, V4) are known to encode low-level visual features (e.g., orientation, colour, spatial frequency and edges), whereas the down-steam visual areas process higher-level visual and semantic features (e.g., object identity, object category, scene location) \cite{FellEsse1991, DiCaCox2007, EpstBake2019}.

There are several brain regions in the human cortex that are specialized in processing of higher-level visual information, such as the lateral occipital cortex (LOC), fusiform face area (FFA), parahippocampal place area (PPA), retrosplenial complex (RSC) and occipital place area (OPA) \cite{KanwMcDeChun1997, GrilKourKanw2001, MalcGroeBake2016, GroeSilsBake2017, EpstBake2019}. 

These brain regions play different roles in supporting object and scene perception. For example, LOC and FFA are located in the inferior temporal (IT) cortex, and are specialised in object and face recognition, respectively \cite{KanwMcDeChun1997, GrilKourKanw2001}. On the other hand, PPA, RSC, and OPA encode different aspects of scene processing \cite{MalcGroeBake2016, GroeSilsBake2017, EpstBake2019}. Specifically, PPA is more sensitive to local spatial features of the scene, while OPA is more involved in large-scale visual features \cite{WatsAndrHart2017}. By contrast, RSC is involved in more abstract scene representations and navigation in the wider environmental space, with less direct associations to image features \cite{WatsAndrHart2017}. 
It remains unclear what types of computations these regions are performing, and how do they fit in the context of larger brain networks. There is also increasing evidence suggesting that each region is sensitive to multiple scene properties that may be correlated, and that their neural representations are likely to depend on the tasks or behavioural goals of the observer \cite{MalcGroeBake2016}.

The striking similarity between the architecture of deep neural networks and the hierarchical structure of the human visual system has made deep learning outperform other recently developed models developed at explaining neural activity underlying object and scene perception \cite{KhalKrie2014, CichKhosPant+2016, HoriKami2017, YamiHongCadi+2014, EickGramVaroThir2017, GuclGerv2015}.

One influential study combining neuroimaging methods and deep learning has recently compared the stage-wise representations of visual objects in an artificial deep neural network (DNN) to the temporal and spatial representations in neural activity measured via magnetoencephalography (MEG) and functional magnetic resonance imaging (fMRI), respectively \cite{CichKhosPant+2016}. They found the hierarchical relationship between the representations in the DNN and neural data. Specifically, the earlier layers of the DNN represent object-based information that are more similar to those encoded in the low-level visual areas, while the DNN representations at the later layers better match the neural representations in the higher-level visual areas. In addition, they found that these matched representations between the DNN and neural data evolved rapidly over time and they emerged in sequence from the early to later layers, mimicking feedforward connections along the visual hierarchy \cite{FellEsse1991, ZoccKouhPoggDiCa2007}. 

Moreover, a recent study has discovered that recurrent neural network could outperform feedforward deep learning models at predicting late neural activity in the non-human primate IT cortex during object recognition, emphasising the important role of feedback connections between cortical areas at untangling high-level visual information \cite{KarKubiSchm+2019}. Further elucidation of the network and functional organisation of the brain regions involved in visual perception could provide guidance for designing deep generative models that can learn more complex compositional representations for visual processing tasks \cite{Krie2015}.

In addition to the ventral visual pathway, selective visual information processing also involves the dorsal frontoparietal attention network, thought to play a central role in enhancing task-relevant sensory information while filtering out irrelevant information \cite{SereYant2006, BuscKast2015, ScolSeidKast2015}. There are multiple neural mechanisms thought to support the disentangling of relevant and irrelevant sensory information. These include sensory gain, neuronal noise modulation, and selective pooling mechanisms \cite{HillVogeLuck1998, MartTreu2002, ReynPastDesi2000, PestCarrHeegGard2011, ItthGarcRung+2014, ItthChaByerSere2017, ItthSere2016, ReynHeeg2009, MitcSundReyn2009, CoheMaun2009}. These neural computations allow attention to assign different weights to competing visual inputs across different spatial locations and across low- and higher-level visual features. 


In sum, deep learning models can serve as a useful computational model for human visual perception. CNNs have already proved useful as a computation model of the primate visual system for simple tasks such as object recognition \cite{GuclGerv2015, CichKhosPant+2016, RajaIssaBash+2018, KhalKrie2014, HoriKami2017, YamiHongCadi+2014, EickGramVaroThir2017}. 
Deep learning models that can learn rich and meaningful disentangled representations such as our SPLIT framework could be useful for further studying the disentangling neural mechanism in the brain.


\section{Related Work}

The topic of learning disentangled representations has received much attention recently, particularly in the context of latent variable models such as VAEs. Efforts in this direction can be classified into two approaches.
The first approach focuses on the effect of regularisation by dissecting the objective function and proposing a modified version with regularisation terms that emphasise different aspects of the latent structures and features of the learnt representations \cite{hoffman2016elbo,chen2018isolating,esmaeili2018hierarchical, chen2016infogan}. We can view this family of methods as disentanglement using biased objective functions. However, there is only a limited control over what features or structures are learnt in the latent representations. Moreover, without augmenting the model with more complex structure, this approach is limited to learning simple representations.

Another approach is to bias the model structure by explicitly imposing latent structure through choices of architecture of the encoders and/or decoders such that specific latent variables have their specific roles in the representations \cite{dilokthanakul2016deep, eslami2016attend, crawford2019spatially, NIPS2019_8387, van2016conditional, nguyen2019hologan, greff2016tagger}. These methods enforce strong and potentially more complex structural biases on the latent representations. Since the disentanglement is done at the structure of the model, our SPLIT framework is directly applicable to this type of methods. 

We note that the two approaches overlap. Adding regularisers to the objective can often be viewed as modifications to the model structure \cite{MathRainSiddTeh2019}. Conversely, imposing a complex latent representation structure adds additional terms to the objective. 

The closely related work by Zhao et al.\cite{zhao2017learning} tackles the problem of multi-scale disentanglement with VAEs. In their work, disentanglement is achieved through a careful architectural design, where more abstract information goes through more computational layers.
Similarly, in the implicit autoencoder \cite{makhzani2018implicit}, local and global information can be controlled via the number of latent dimensions and noise vector. Importantly however, these ways of imposing inductive biases require an iterative design process of the model structure. For instance, the number of layers needs to be tuned by observing the resulting representations after training. This iterative design requirement makes it less scalable and difficult to extend the model, e.g. by adding additional independent structural biases. Jakab et al. \cite{jakab2018unsupervised} introduce an architecture that can learn key-points from pairs of source and target images. The architecture resembles our SPLIT framework with two sets of latent variables, one for key-points (global geometric representation) and another for image style. But unlike our work, they encourage the model to represent key-points by using a \emph{spatial softmax} operator to bottleneck the information. 

Unlike other approaches that impose structural biases in the models, our SPLIT framework does not require an extensive modification to the original model architecture. This makes our approach generic. It is applicable to VAEs with any prior structure including richly structured models such as SPAIR. The SPLIT framework can also be viewed as a bias imposition method via modified objective function. However, the additional terms in the objective and the gradients come from an additional VAE that models auxiliary data $\hat{x}$. This idea resembles how a generative adversarial network (GAN) \cite{goodfellow2014generative} uses another deep neural network, called the discriminator, to provide a learnable objective for the main network.  

Finally, there is a parallel direction attempting to solve the problem of spurious correlations in the supervised learning setting. Wang et al. \cite{NIPS2019_9237} have shown that by regularising the predictive power of early convolutional layers, they can achieve a more robust global representation, making their model insensitive to variations in the local features and improving supervised learning score in out-of-distribution test data.  

\section{Conclusion}

We propose a simple extension to the VAE framework that explicitly imposes an inductive bias to encourage the disentanglement into global and local representation. 

We demonstrated the utility of our framework in three VAE models: vanilla VAE, GMVAE and SPAIR. For the vanilla VAE, we investigated the disentanglement quality by (i) visual inspection, (ii) investigation of generative samples using a pre-trained classifier and (iii) a style transfer task. The results illustrated that the SPLIT-VAE can effectively perform the local-global disentanglement.

For the GMVAE, we impose an inductive bias on an unsupervised clustering task by encouraging clusters to only consider the global information. We investigated the clustering representation with (i) visual inspection of the generated samples, (ii) visual inspection of the encoded unseen samples and (iii) a SVHN unsupervised clustering task. We found that SPLIT-GMVAE can cluster data by effectively ignoring local information and achieved better scores than GMVAE in the SVHN clustering benchmark.
The discrete representation was easier to interpret than a vanilla GMVAE.
For example, SVHN images were clustered into groups by digit identity, and the CelebA data were grouped by gender or face orientation information.  

For SPAIR, we used the SPLIT framework to impose an inductive bias on the object detection part of the model to encourage the model to search for objects while ignoring local features. We found that such an inductive bias improved the learning speed of the unsupervised object detection model and resulted in significantly improved object counting scores.

For future work, we hope to see how the ability to disentangle local and global information can lead to improvements in other machine learning tasks. For example, a classifier model could take advantage of local-global disentangled representations by focusing on more relevant features, making the model potentially more robust to adversarial examples.
Another example are visual robot manipulation tasks, where it would be beneficial if the agent could learn locally invariant representations and ignore local information such as colour, texture or lighting in objects. This would result in better generalisation and faster learning in the agent.

\ifCLASSOPTIONcompsoc
  \section*{Acknowledgments}
\else
  \section*{Acknowledgment}
\fi

This research is based upon work supported in part by the Thailand Science Research and Innovation (TSRI) via SRI62W1501. 
Nat Dilokthanakul and Rujikorn Chanrakorn are the main contributors and contributes equally to this work. The authors would like to thank Murray Shanahan for discussions on the preliminary ideas and results, which built the foundation for this work.
The authors also would like to thank Supasorn Suwajanakorn, Suttisak Wizadwongsa, Pitchaporn Rewatbowornwong, Theerawit Wilaiprasitporn, Supanida Hompoonsap, Sarana Nutanong and other members at IST, VISTEC, for useful discussions and supports. 
\ifCLASSOPTIONcaptionsoff
  \newpage
\fi



\bibliographystyle{IEEEtran}
\bibliography{main.bib}
%

%

\begin{IEEEbiography}[{\includegraphics[width=1in,height=1.25in,clip,keepaspectratio]{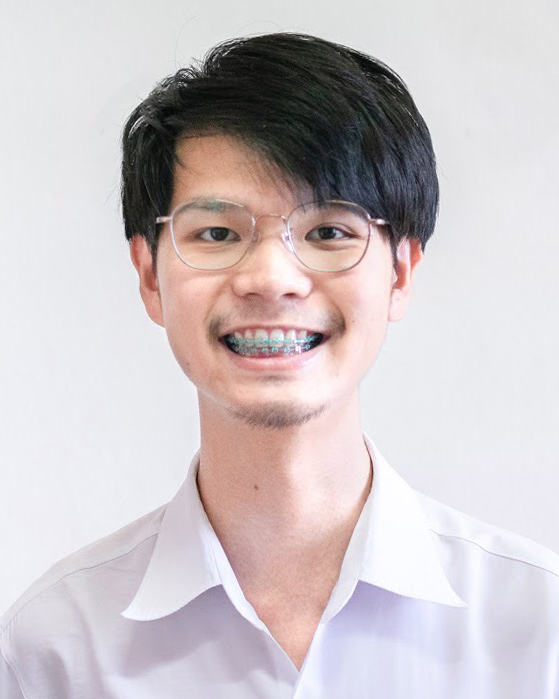}}]{Rujikorn Charakorn}
received a B.Eng. in Computer Engineering from Chulalongkorn Universitiy. He is currently a Ph.D student at the School of Information Science and Technology, Vidyasirimedhi Institute of Science and Technology (VISTEC). His research interests include deep reinforcemenet learning, cooperative and competitive multi-agent learning and computer vision.
\end{IEEEbiography}
\begin{IEEEbiography}[{\includegraphics[width=1in,height=1.25in,clip,keepaspectratio]{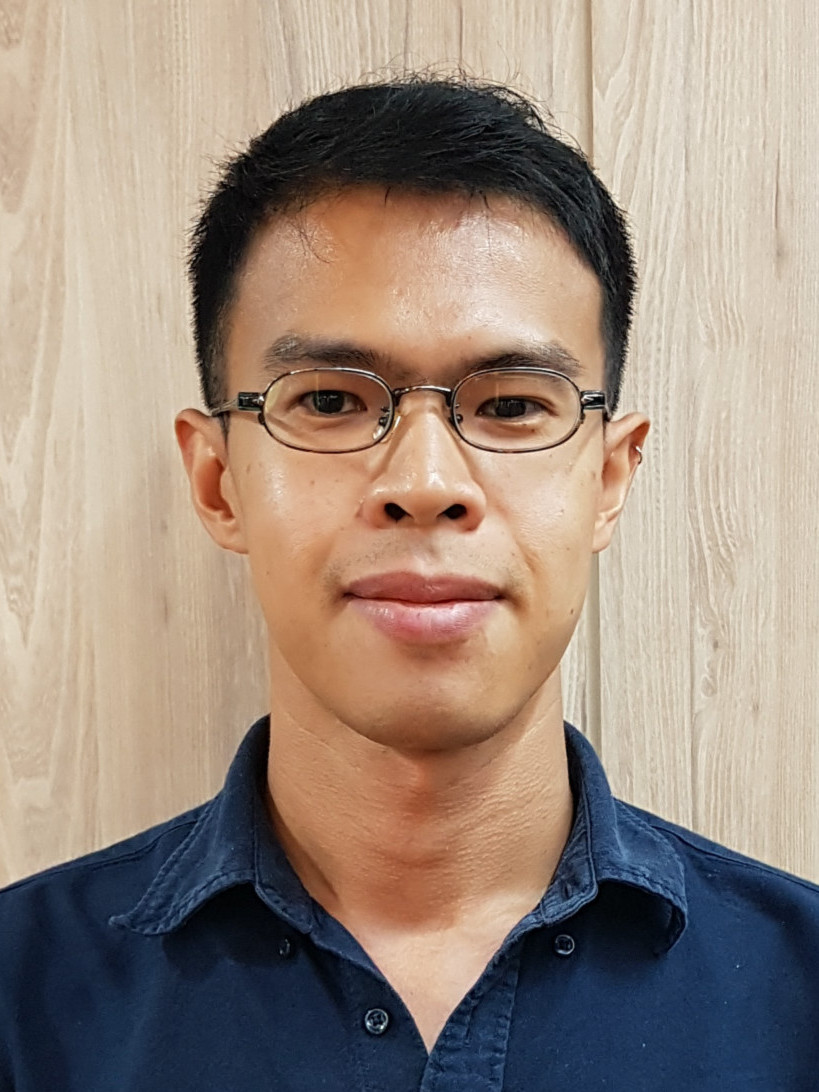}}]{Yuttapong Thawornwattana}
received the MSc in Computational Statistics and Machine Learning from University College London in 2013. He is a researcher in the School of Information Science and Technology, Vidyasirimedhi Institute of Science and Technology (VISTEC). His research interests include Bayesian modelling and statistical machine learning.
\end{IEEEbiography}

\begin{IEEEbiography}[{\includegraphics[width=1in,height=1.25in,clip,keepaspectratio]{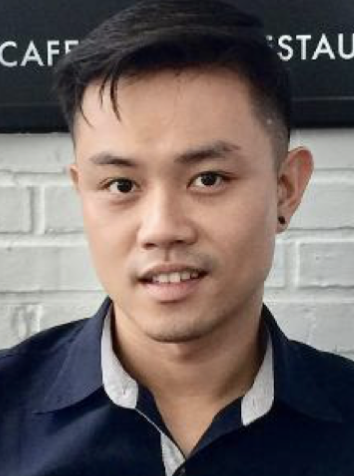}}]{Sirawaj Itthipuripat} received BS in Neuroscience and Psychology from Duke University (2011) and MS/PhD in Neurosciences from University California, San Diego (2013/2017). He is currently a researcher at Learning Institute, King Mongkut's University of Technology Thonburi (KMUTT). He studies neural mechanisms that support attention, learning, and memory using electrophysioloy, neuroimaging, and computational methods.
\end{IEEEbiography}

\begin{IEEEbiography}[{\includegraphics[width=1in,height=1.25in,clip,keepaspectratio]{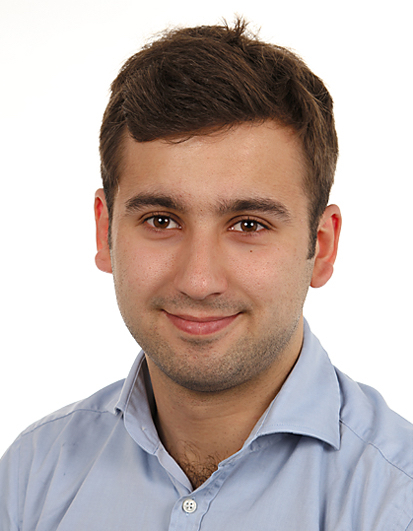}}]{Nick Pawlowski}
received the MSc in Artificial Intelligence from the University of Edinburgh in 2016. He is currently pursuing a Ph.D. in Computing at Imperial College London supervised by Ben Glocker in the field of machine learning with applications to medical image analysis.
\end{IEEEbiography}

\begin{IEEEbiography}[{\includegraphics[width=1in,height=1.25in,clip,keepaspectratio]{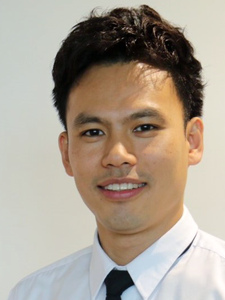}}]{Poramate Manoonpong}
is a Professor with the School of Information
Science and Technology, VISTEC, Thailand, and an Associate
Professor for embodied AI and robotics with the University of Southern
Denmark (SDU), Odense, Denmark. 
He is also a Professor with the
College of Mechanical and Electrical Engineering, Nanjing University of
Aeronautics and Astronautics (NUAA), Nanjing, China. He researches embodied AI, neural locomotion control and robotics. 
\end{IEEEbiography}
\begin{IEEEbiography}[{\includegraphics[width=1in,height=1.25in,clip,keepaspectratio]{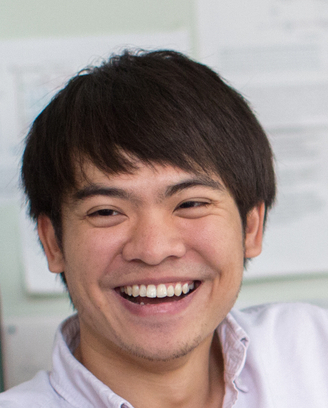}}]{Nat Dilokthanakul}
received the MSci in
Physics from University College London in 2014 and the PhD degree from Imperial College London, in 2019. He is currently a postdoctoral researcher in the School of Information Science and Technology, Vidyasirimedhi Institute of Science and Technology (VISTEC). His research interests include representation learning, deep reinforcement learning and generalisation in machine learning.
\end{IEEEbiography}
\vfill






\end{document}